\documentclass[11pt]{article}              

\usepackage[margin=25mm]{geometry}
\usepackage{subfig}
\usepackage{graphicx}
\usepackage{colortbl}
\usepackage{fancyhdr}
\usepackage{amssymb,amsfonts,amsmath,amsthm}
\usepackage{natbib}
\usepackage{bstnotations}

\usepackage{authblk}
	
\graphicspath{{./},{./Figures/}}

\begin{document}
\title{Learning non-stationary and discontinuous functions using clustering, classification and Gaussian process modelling} 

\author[1]{M. Moustapha} \author[1]{B. Sudret}

\affil[1]{Chair of Risk, Safety and Uncertainty Quantification,
  
  ETH Zurich, Stefano-Franscini-Platz 5, 8093 Zurich, Switzerland}

\date{}
\maketitle

\abstract{Surrogate models have shown to be an extremely efficient aid in solving engineering problems that require repeated evaluations of an expensive computational model. They are built by sparsely evaluating the costly original model and have provided a way to solve otherwise intractable problems. A crucial aspect in surrogate modelling is the assumption of smoothness and regularity of the model to approximate. This assumption is however not always met in reality. For instance in civil or mechanical engineering, some models may present discontinuities or non-smoothness \emph{e.g.}, in case of instability patterns such as buckling or snap-through. Building a single surrogate model capable of accounting for these fundamentally different behaviours or discontinuities is not an easy task. In this paper, we propose a three-stage approach for the approximation of non-smooth functions which combines clustering, classification and regression. The idea is to split the space following the localized behaviors or regimes of the system and build local surrogates that are eventually assembled. A sequence of well-known machine learning techniques are used: Dirichlet  process mixtures models (DPMM), support vector machines and Gaussian process modelling. The approach is tested and validated on two analytical functions and a finite element model of a tensile membrane structure.\\[1em] 

}

\maketitle

\section{Introduction}
Computational models, which allow scientists and engineers to accurately simulate complex systems and predict their behaviour in various contexts, are nowadays  a key tool present in virtually all fields of applied sciences and engineering. Cast as computer experiments, they are able to predict with high fidelity the behaviour of the studied system in replacement of, or as a complement to laboratory experiments. The downside of such high-fidelity models is however that they are computationally demanding. This is even more relevant in the context of uncertainty quantification or design optimization, where the models need to be evaluated multiple times. 

Surrogate models have become paramount in such fields as they allow for an efficient solution of otherwise computationally intractable problems. They are inexpensive proxies that can be used \emph{in lieu} of expensive computational models. Examples of such surrogates include Gaussian process models also known as Kriging \citep{Santner2003, Rasmussen2006}, polynomial chaos expansions \citep{Xiu2002,BlatmanJCP2011}, support vector machines \citep{Vapnik:1995}, polynomial response surfaces \citep{Myers2002}, etc. These methods have been applied in various problems pertaining to uncertainty quantification or design optimization. The use of surrogate models in such fields are now mature as shown by the recent reviews in reliability analysis \citep{Teixeira2021,MoustaphaSS2022}, Bayesian inversion \citep{Yan2017} or design optimization \citep{Chatterjee2019,MoustaphaSMO2019}.

In most of these applications, it is assumed that the computational models to approximate feature some accommodating properties such as smoothness, differentiability or stationarity. Yet there exists cases when these assumptions do not hold. In mechanical engineering, this may happen for instance when solving non-linear problems involving instability such as snap-through or bifurcations in the solution path, \emph{e.g.}, crash simulation. In computational fluid dynamics, simulations of compressive flows that involve shocks also belong to this category. In other cases, the underlying phenomenon may present different localized features or extreme regime variations which are strongly dependent on the inputs.

Various methods have been developed in the field of uncertainty quantification to tackle such problems. The first class of methods borrows from digital signal processing and image detection to identify discontinuities or strong gradients of the function to approximate using techniques such as polynomial annihilation \citep{LeMaitre2004b,Gorodetsky2012}. Such approaches however rely on uniformly sampled grids and are often limited to two-dimensional problems. \citet{Sargsyan2012} proposed a technique combining Bayesian inference and polynomial chaos expansions that does not require using a regular grid and hence allowing for a reduced number of samples. However, their approach was also developed for two-dimensional problems and the authors did not investigate how well it scales with dimensionality. 

Another class of methods relies on Gaussian process (GP) regression where the irregularities on the model to approximate are tackled by introducing non-stationary covariance functions or kernels. Indeed, such kernels allow one to capture heterogeneous variations or heteroscedastic noise while keeping the computational budget low. The direct approach to build such kernels is to consider the noise  variance, signal variance and/or characteristic length scale to be input-dependent, such as in \citet{Paciorek2003}. \citet{Heinonen2016} proposed an approach where all three  parameters are considered latent variables and inferred as hyper-parameters of the GP. Such an approach has shown increased efficiency compared to vanilla GP but it also comes with an increased inference cost due to the fact that there are no more closed-form solution and the hyperparameters need to be calibrated using sampling based techniques (See \citet{Rasmussen2002}). Furthermore, they do not allow to tackle problems with discontinuities. 

A more sensible approach based on non-stationary GP consists in splitting the input space using for instance treed Gaussian processes or a mixture of experts \citep{Tresp2000,Rasmussen2002,Meeds2005}. Similarly, it is also possible to define non-stationary Gaussian process models by partitioning the training data into smaller subsets using clustering techniques, such as in \citet{ZhangYiming2019} and \citet{Konomi2019}, where K-means and nearest-neighbors clustering are used. Such approaches also have the advantage of offering faster training and testing of the model as the experimental design is divided into smaller and more computationally manageable subsets.  Finally, another popular way to define non-stationary kernels is by warping the input, and sometimes the output, space. By doing so, one may find a latent space where the function to approximate is smoother. Examples of such techniques include warped GP \citep{MarminThesis2018} or manifold GP regression \citep{Calandra2016, Kuleshov2018}.

In this work, we will focus on multi-stage techniques where the problem is solved by using a sequence of well-known machine learning techniques. More specifically, we consider the class of methods based on the following three-stage approach: clustering, classification and regression \citep{Boroson2017,Dupuis2018}. \citet{Basudhar2008a,Serna2009} were the first to propose decomposing the problem of identifying multiple failure domains of mechanical systems using support vector machines. However, they do not include the regression step as they are only concerned with an optimization problem where only the state of a sample is of interest (\emph{i.e.}, whether it belongs to the failure domain or not). \citet{MoustaphaThesis2016,MoustaphaUNCECOMP2019} extended the approach to the prediction of the model responses by building local Kriging surrogates in each identified domain. However in all these approaches, it was assumed that the clusters were identified either using expert knowledge or by only considering the model responses which span different ranges. \citet{Niutta2018} proposed identifying the clusters by detecting jumps in the model responses for relatively close samples. However, this technique works only in low-dimensional problems and when the response of different clusters are disjoint. This is a strong limitation and was to some extent overcome by using joint clustering of both the inputs and outputs in \citet{Bernholdt2019}. In that work, they use K-means clustering to identify the clusters and multi-layer perceptrons for classification and regression tasks. The number of clusters is defined here using the elbow approach, which is a visual technique requiring user interaction. Furthermore it is not robust w.r.t. the initialization of the K-means algorithm and noise in the data. More generally, an important limitation in the contributions presented above is that the three steps are disconnected and the prediction uncertainty in one step is not accounted for in the subsequent ones. 

In this paper, we propose an approach that aims at solving these two limitations. First, to automatically identify the number of clusters in a robust way, we consider a non-parametric Bayesian technique, namely \emph{Dirichlet process mixture models} (DPMM). The interest in using DPMM are three-fold: i. they automatically estimate the optimal number of clusters according to patterns identified in the data, ii. they offer a probabilistic framework that allows one to propagate the epistemic uncertainty related to this clustering task to both the subsequent classification and regression steps, and iii. they are flexible enough and their complexity can grow as new data is observed (for instance in an active learning scheme, where new regimes of the model could be identified).

In the remainder of this paper, we first present the three-stage methodology and how the steps are connected in Section~\ref{sec:setup}. In Section~\ref{sec:Methods}, we present in details the three methods used in each step, namely, Dirichlet process mixture models, support vector machines for classification and Gaussian process modelling. We finally illustrate the proposed approach in Section~\ref{sec:Examples} using two analytical examples and an engineering application related to the design of a tensile membrane structure \citep{Valdes-Vazquez2020,Valdes-Vazquez2021}.

\section{Problem set-up and three-stage approach}\label{sec:setup}
Let us consider a set of $N$ data points $\prt{\mathcal{X},\mathcal{Y}}$ where $ \mathcal{X} = \acc{\ve{x}^{(i)} \in \mathbb{X} \subset \mathbb{R}^M, i = 1, \ldots N}$ is a set of $M$-dimensional inputs and $\mathcal{Y}$ are corresponding scalar outputs such that \\$\mathcal{Y} = \acc{y^{(i)} = \mathcal{M}\prt{\ve{x}^{(i)}} \in \mathbb{R}, i = 1 \ldots N }$. The model $\mathcal{M}$ is assumed black-box, meaning that it is only accessible through an evaluation over a finite set of input points. We further assume in this setting that the model is non-smooth, \emph{i.e.}, it exhibits sharp localized features and, most noticeably, discontinuities. As the model can only be evaluated on a finite set of samples, discontinuities in the current work is assumed when the model presents extreme variations in the outputs for seemingly close input points. 

The goal of the analysis is to learn the input-output relationship of the model $\mathcal{M}$ through the limited set of training data $\mathcal{D} = \prt{\mathcal{X},\mathcal{Y}}$, also known as \emph{experimental design}. This ultimately leads to a cheaper-to-evaluate surrogate model that can be used to predict the response of the model for any new point. Generally, this type of problems is tackled using regression techniques where a class of parameterized models are assumed and then their hyper-parameters are calibrated so as to minimize a generalization error. Such models would however fail when there are discontinuities or heterogeneous variations associated to limited observations.

In this work, we consider tackling this problem by splitting the space along the discontinuities and building local regression models in each of the obtained subdomains. To achieve this, we consider a three-stage framework which is illustrated in Figure~\ref{fig:Illus} and summarized as follows:
\begin{figure}[!ht]
	\centering
	\includegraphics[width=0.7\textwidth]{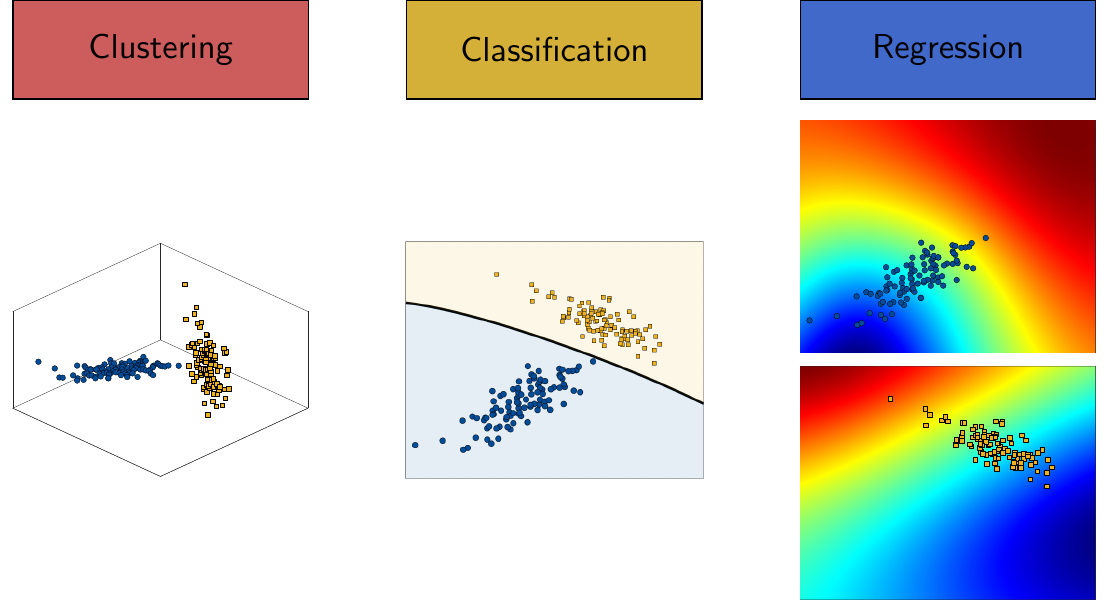}%
	\caption{Illustration of the three-stage approach.}
	\label{fig:Illus}
\end{figure}
\begin{enumerate}
	\item \textbf{Clustering:} The first learning step aims at identifying patterns in the data that hint to subdomains separated by discontinuities. To achieve this, we cluster the \emph{joint input-output} data points. This is an unsupervised learning problem for which numerous techniques have been developed \citep{Pham2017}. $K$-means clustering \citep{Lloyd1982} is probably the most common approach thanks to its simplicity. However, it assumes that the number of clusters is known and further fails when the clusters are of disproportionate sizes. Another approach that partially overcomes difficulties related to $K$-means clustering are Gaussian mixture models which offer a probabilistic framework for clustering \citep{Rokach2005}. They hence allow for a more nuanced clustering of the data by providing soft cluster memberships, \emph{i.e.}, each data point is assigned with a probability of belonging to a given cluster. This feature allows one to solve more complex problems, \emph{e.g.}, when the clusters are partially overlapping. However, similarly to $K$-means, they assume that the number of clusters is known in advance. In general, trial-and-errors approaches are used to define the optimal number of clusters for such problems, which is not optimal. 
	
	We therefore consider in this work a more holistic approach where the number of clusters is also inferred from the data using a non-parametric Bayesian model, more specifically Dirichlet process mixture models \citep{Li2019} as described in Section~\ref{sec:DPM}. 
	
	At the end of this step, the experimental design is split into $K$ subsets $\mathcal{C}_k, \, k = 1, \ldots, K$. 
	\item \textbf{Classification:} Assuming that the data have been clustered, we can now place labels on them and turn to supervised learning. More specifically, let us assume $K$ clusters are identified in the previous step. We thus define the labels $\acc{\ell_1 \enum \ell_K}$ and the labelled training data $\mathcal{X} \times \mathcal{L}$ where each couple $\prt{\ve{x}^{(i)}, \ell^{(i)}}$ is defined such that $\ell^{(i)} = \ell_k$ if $\prt{\ve{x}^{(i)}, y^{(i)}} \in \mathcal{C}_k$. The goal of this step is then to partition the input space such that any new sample can be mapped to at least one of the clusters $\mathcal{C}_k$. This will ultimately allow us to select the appropriate local regression model(s) to evaluate the new point.
	
	This task is carried out in this work by using support vector machines (SVM) for binary and multi-class classification \citep{Vapnik:1995}. The probabilistic framework is introduced by considering Platt's approach to computing posterior probabilities given a binary SVM prediction \citep{Platt2000}. For multi-class problems, binary classifiers are appropriately combined to provide both class membership and posterior probabilities.
	\item \textbf{Regression:} In this final step, Gaussian process (GP) models \citep{Rasmussen2006} are employed to make the final prediction. We further investigate the use of three different approaches for combining the various GP models built in this stage. In the first two approaches, local surrogate models $\widehat{\mathcal{M}}_k$ are built for each of the $K$ identified clusters. When it comes to prediction, the recombination is made as follows:
	\begin{itemize}
		\item \emph{Hard recombination}: In this approach, the surrogate model which corresponds to the cluster predicted by the classifier is solely used to make the final prediction, \emph{i.e.},
		\begin{equation}
			\widehat{\mathcal{M}}\prt{\ve{x}} = \sum_{k=1}^K \mathbbm{1}_{\mathcal{C}_k}\prt{\ve{x}} \widehat{\mathcal{M}}_k \prt{\ve{x}},
		\end{equation}
	where $\mathbbm{1}_{\mathcal{C}_k}\prt{\ve{x}}$ is equal to $1$ if $\ve{x}$ is predicted to belong to the cluster $\mathcal{C}_k$, \emph{i.e.},  $\mathcal{M}^{\text{SVC}}\prt{\ve{x}} = \ell_k$ and $0$ otherwise;
	\item \emph{Soft recombination}: In this approach, the prediction for each point is obtained as a weighted combination of all the local surrogate models, \emph{i.e.},
		\begin{equation}
	\widehat{\mathcal{M}}\prt{\ve{x}} = \sum_{k=1}^K w_k\prt{\ve{x}} \widehat{\mathcal{M}}_k \prt{\ve{x}},
	\end{equation}	
	where the weight $w_k\prt{\ve{x}} \in \bra{0,\,1}$ with $\sum_{k=1}^K w_k\prt{\ve{x}} = 1$ may be related to the actual probability that the point $\ve{x}$ belongs to the cluster $\mathcal{C}_k$ as defined by the classifier. 
	\item \emph{Categorical recombination}: Contrary to the previous two approaches, a single Gaussian process model is built here. This is achieved by using an additional variable which is a categorical parameter indicating which cluster a given point belongs to, \emph{i.e.}, the training set is the couple $\acc{\mathcal{X}, \mathcal{L}} \times \mathcal{Y}$ where $\mathcal{L} = \acc{\ell^{(i)}, i = 1 \enum N}$ are the labels of the training set identified in the clustering stage. The surrogate model is therefore built on a space of dimension $M+1$: $\widehat{\mathcal{M}}\prt{\ve{x}} = \widehat{\mathcal{M}}^{\text{cat}}\prt{\widetilde{\ve{x}} = \prt{\ve{x}, \widehat{\ell}\prt{\ve{x}} }}$, where the categorical variable is given by the SVC prediction, \emph{i.e.}, $\widehat{\ell}\prt{\ve{x}} = \mathcal{M}^{\text{SVC}}\prt{\ve{x}}$.
	\end{itemize} 
\end{enumerate}

The following section describes in details each of the ingredients introduced in the proposed framework.

\section{Description of the components of the proposed method}\label{sec:Methods}
\subsection{Clustering using Dirichlet process mixture models}\label{sec:DPM}

\paragraph{Gaussian mixture models\\}
Let us now consider the set of available data $\mathcal{W} = \acc{\ve{w}^{(i)}, i = 1 \enum N}$, where $\ve{w}^{(i)}  = \prt{\ve{x}^{(i)}, \, y^{(i)}}$ is a vector gathering both inputs and outputs, and let us assume that they are associated to some latent variables $\ve{z}$. In a clustering set-up, say using a Gaussian mixture, the latent variables would be $\ve{z} = \acc{\ve{\pi},\ve{\mu},\ve{\Sigma}}$ where $\ve{\pi}$ are mixing coefficients and $\ve{\mu}$ and $\ve{\Sigma}$ are the mean and covariance of multivariate normal random variables. The goal is then to find the posterior distribution $p(\ve{z}|\ve{w})$ of the latent variables given the data and using Bayes rules, \emph{i.e.},
\begin{equation}
	p(\ve{z}|\ve{w}) = \frac{p\prt{\ve{w},\ve{z}}}{p\prt{\ve{w}}} = \frac{p\prt{\ve{w}|\ve{z}}p\prt{\ve{z}}}{p\prt{\ve{w}}} \propto p\prt{\ve{w}|\ve{z}}p\prt{\ve{z}},
\end{equation}
where $p\prt{\ve{w}|\ve{z}}$ is the data likelihood, $p\prt{\ve{z}} = p\prt{\ve{\pi},\ve{\mu},\ve{\Sigma}}$ is the prior over the latent variables and $p\prt{\ve{w}}$ is the evidence. 

The prior can be fully factorized into $p\prt{\ve{\pi}}p\prt{\ve{\mu}}p\prt{\ve{\Sigma}}$ since the three parameters are considered mutually independent. The prior on the mixing coefficients $p\prt{\ve{\pi}}$ is usually chosen as a Dirichlet distribution with parameters $\alpha/K$ where $\alpha$ is a positive scaling parameter and $K$ is the predefined number of clusters:
\begin{equation}
	p\prt{\pi_1,\ldots, \pi_K|\alpha} = \textrm{Dirichlet}\prt{\alpha/K,\ldots, \alpha/K} = \frac{\Gamma \prt{\alpha}}{\Gamma\prt{\alpha/K}^K}\prod_{k=1}^{K}\pi_k^{\alpha/K-1},
\end{equation}
where $\Gamma$ is the Gamma function.

The Dirichlet distribution is chosen precisely because it is the {\em conjugate distribution} to the multinomial distribution, which is used for clusters membership assignment, later denoted by $c$. The generative model for data derived from a Gaussian mixture model can therefore be cast as
\begin{equation}\label{eq:GMM}
	\begin{split}
		\pi_k & \sim \textrm{Dirichlet}\prt{\alpha/K,\ldots, \alpha/K}, \quad k = \acc{1, \ldots, K}, \\
		c^{(i)}  & \sim \textrm{Multinomial}\prt{\pi_1, \ldots, \pi_K}, \quad i = \acc{1, \ldots, N}, \\
		\ve{w}^{(i)} | \acc{c^{(i)} = k} & \sim \mathcal{N}\prt{\ve{\mu}_k, \ve{\Sigma}_k}, \quad \qquad  \qquad \qquad  i = \acc{1, \ldots, N},
	\end{split}
\end{equation}
where $\ve{\mu}_k$ and $\ve{\Sigma}_k$ are respectively the mean and covariance parameters of each local Gaussian distribution in the mixture.

It is generally assumed in such a model that $K << N$, which in other words means that samples from all clusters have been observed. However, there may exist cases when $K$ is in the same order or even larger than $N$. An alternative view to such cases is that at any moment all clusters have not yet been observed and drawing more data from the generative model will reveal new clusters. This naturally leads to extending this finite mixture model into an infinite one using non-parametric Bayesian models whose complexity can grow as more data are observed. 
 
This is precisely what a Dirichlet process mixture model does. It generalizes the generative model described in Eq.~\eqref{eq:GMM} by assuming an infinite number of clusters, \emph{i.e.}, that $K \rightarrow \infty$. This corresponds to choosing a Dirichlet process \citep{Ferguson1973} as prior for the mixing coefficients, as explained in the sequel.

\paragraph{Dirichlet process\\}
A Dirichlet process (DP) is a distribution over distributions defined by a base distribution $G_0$ and a positive scaling parameter $\alpha$. The output from a Dirichlet process is therefore a discrete distribution. It is however not possible to directly draw from $G$ considering the formal definition of a Dirichlet process. Other alternative views such as the Chinese restaurant process \citep{Aldous1985}, the P\'olya urn scheme \citep{Blackwell1973} or the stick-breaking representation \citep{Sethuraman1994} have been proposed instead. 

\begin{figure}[!ht]
	\centering
	\includegraphics[width=0.5\textwidth]{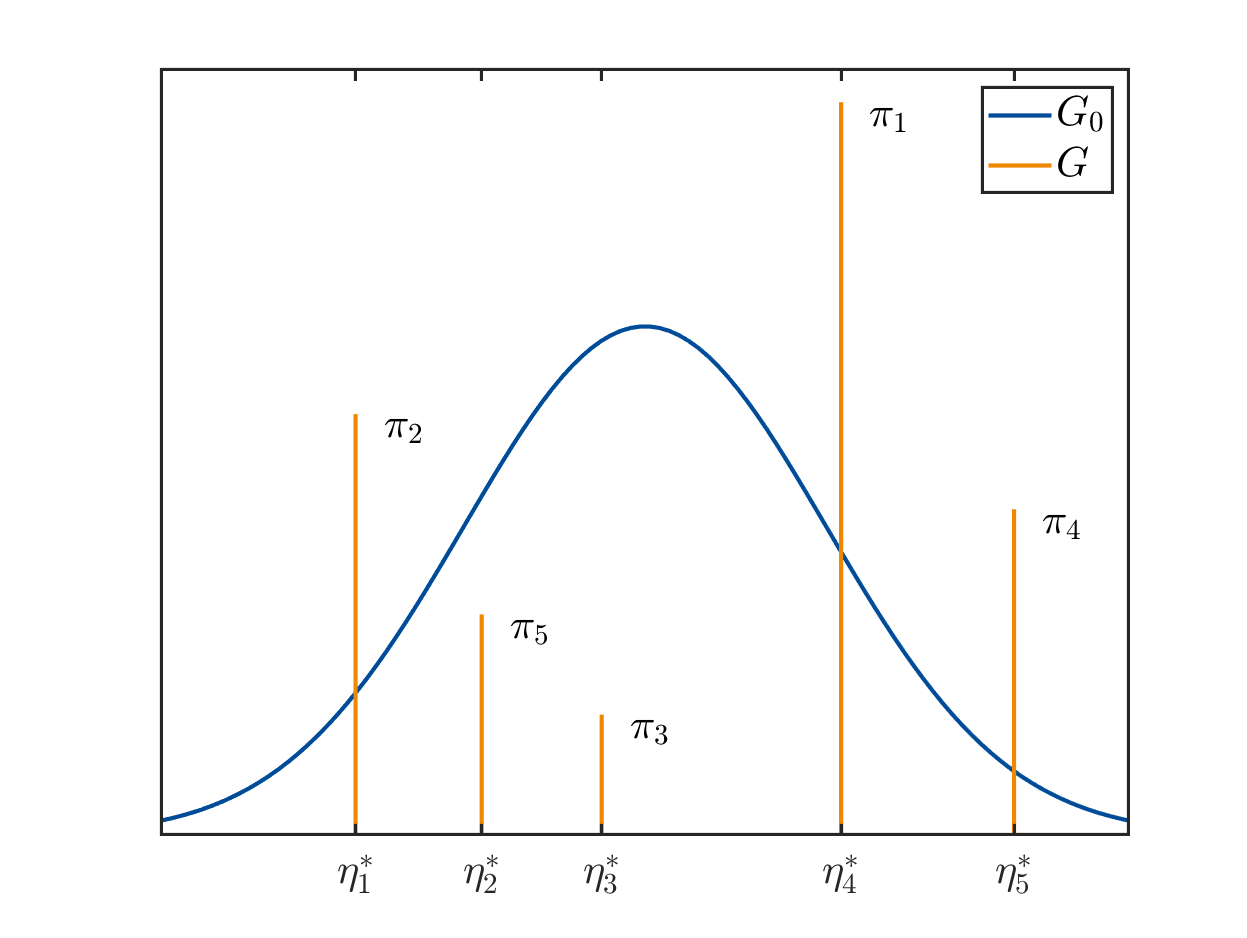}%
	\caption{Illustration of a Dirichlet process: $G_0$ is the base distribution from which the atoms $\eta^\ast_k$ are sampled, $\pi_k$ are the corresponding weights and $G$ a realization of the DP.}
	\label{fig:Illus:DP}
\end{figure}

In this work, we consider the latter approach. More specifically, let us consider an infinite collection of two random variables $V_k \sim \textrm{Beta}(1,\alpha)$ and $\eta_k^\ast \sim G_0$ with $k = \acc{1,2, \ldots}$. The stick-breaking representation of $G$ is then defined as follows:
\begin{equation}
	\begin{split}
	\pi_k & = v_k \prod_{j = 1}^{k-1}\prt{1-v_j}, \\
	G & = \sum_{k=1}^{\infty} \pi_k\prt{\ve{v}} \delta_{\eta_k^\ast}\prt{\eta_k},
	\end{split}
\end{equation}
where $\delta$ is the Kronecker symbol. This representation is illustrated in Figure~\ref{fig:Illus:DP} where the $\eta_k^\ast$ are location parameters also known as atoms and $\pi_k$ are corresponding weights.

In a DP, there is a countably infinite number of atoms and the weights sum up to $1$, making $G$ a discrete distribution.  This infinite set of atoms lends itself to modelling priors in infinite mixture models. More specifically, the DP is used in Dirichlet process mixture models as a non-parametric prior in a hierarchical Bayesian model specified as follows \citep{Antoniak1974,Blei2006}:
\begin{equation}
	\begin{split}
		G|\acc{\alpha,G_0} & \sim DP(\alpha,G_0), \\
		\eta^{(i)} | G & \sim G, \\
		\ve{W}^{(i)} | \eta^{(i)} & \sim p(\ve{w}^{(i)}|\eta^{(i)}).
	\end{split}
\end{equation}
Given a dataset $\mathcal{W}$, each data point $\ve{w}^{(i)}$ is assumed to be generated by first drawing a component label $c^{(i)} = \acc{1, 2, \ldots }$ with probability distribution $p(c^{(i)} = k|\ve{V}) = \pi_k(\ve{v})$ and then drawing  $\ve{w}^{(i)}$ from $p(\ve{w}^{(i)}|\eta_k)$. In this work, $p$ is chosen as a distribution from the exponential family for which $G_0$ is a conjugate prior, which turns out to also belong to the exponential family and hence making inference easier.

\paragraph{Posterior estimation\\}
The latent variables in this setting are therefore $\ve{z} = \acc{\ve{v},\ve{\eta},\ve{c}}$. The goal of the analysis is then to find the posterior distribution of these latent variables given the observed data $\mathcal{W}$, which is denoted by $p\prt{\ve{z}|\mathcal{W}, \ve{\theta}}$. There is no closed-form solution to this problem and typical solution schemes rely on Markov Chain Monte Carlo (MCMC). MCMC algorithms allow one to obtain an approximation of the posterior using Markov chains whose stationary distribution is the sought posterior. The usual approach in Dirichlet process mixture models is Gibbs sampling which is particularly suited to this task as one can have access to the conditional distributions of the latent variables analytically \citep{Neal2000,Ishwaran2001}. However, the difficulty with MCMC algorithms is that they are expensive, as they require a large number of samples, often generated sequentially, and their convergence is difficult to monitor. 

An alternative approach to circumvent these issues is \emph{variational inference}, where the estimation of the posterior is replaced by an optimization problem \citep{Wainwright2003}. More specifically, the intractable posterior is replaced by a parametric family of variation distributions denoted here by $q_\nu(\ve{z}|\nu)$. In this paper, we consider the approach proposed by \citet{Blei2006} which relies on the mean-field approximation, \emph{i.e.}, the variational distribution is fully factorized (all the latent variables are mutually independent). The optimization problem then consists in finding within the selected family of variational distributions the values of the hyperparameters $\nu$ that will minimize the Kullback-Liebler (KL) divergence between the true posterior and its approximation. This quantity reads
\begin{equation}\label{eq:KL}
	\begin{split}
	KL\prt{q_\nu(\ve{z}|\nu)||p\prt{\ve{z}|\mathcal{W}, \ve{\theta}}} & = \int_{-\infty}^{\infty} q_\nu(\ve{z}|\nu) \log\prt{\frac{q_\nu(\ve{z}|\nu)}{p\prt{\ve{z}|\mathcal{W}, \ve{\theta}}}} \textrm{d}\ve{z} \\
	& = \mathbb{E}_{q_{\nu}} \bra{\log q_\nu(\ve{z}|\nu)} -  \mathbb{E}_{q_{\nu}} \bra{ p\prt{\ve{z},\mathcal{W} | \ve{\theta}}}
     + \log p\prt{\ve{w}|\ve{\theta}}.
	\end{split}
\end{equation}
By noting that the divergence is always positive (or using Jensen's inequality), it can be shown that minimizing Eq.~\eqref{eq:KL} is equivalent to maximizing a lower bound of the marginal log likelihood, also referred to as ELBO and denoted by
\begin{equation}
	\log p\prt{\ve{w}|\ve{\theta}} \geq \mathbb{E}_{q_{\nu}} \bra{ p\prt{\ve{z},\mathcal{W} | \ve{\theta}}} - \mathbb{E}_{q_{\nu}} \bra{\log q_\nu(\ve{z}|\nu)}.
\end{equation}
By appropriately choosing the family of variational distributions for each latent variable, it is possible to make the computation of the ELBO tractable. In the approach proposed by \citet{Blei2006} considered here, the factorized variational distribution is cast as
\begin{equation}
	q_\nu(\ve{v},\ve{\eta},\ve{z}|\nu) = \prod_{t=1}^{T-1}q_{\gamma_t}\prt{v_t} \prod_{t=1}^{T}q_{\tau_t}\prt{\eta_t} \prod_{k=1}^{N}q_{\Phi_k}\prt{c_k},  
\end{equation}
where $q_{\gamma_t}\prt{v_t}$ are Beta distributions, $q_{\tau_t}\prt{\eta_t}$ are exponential family distributions and $q_{\Phi_k}\prt{c_k}$ are multinomial distributions. In this equation, the infinite samples is truncated to $T$
terms by setting $q(v_T = 1) = 1$, which implies that $\ve{\pi}_t\prt{\ve{v}} = 0$ for $t \geq T$. The solution to this problem is eventually obtained using a coordinate ascent algorithm for which the incremental updates can be computed analytically \citep{Ghahramani2000}. The reader is referred to \citet{Blei2006} for further details.

\subsection{Classification using support vector machines}\label{sec:SVC}
\subsubsection{Binary classification}
Support vector machines are a popular supervised learning algorithm developed by \citet{Vapnik:1995}. They were developed for binary classification and were later extended to account for multiple classes. Let us first consider the binary case (\emph{i.e.}, assuming only two clusters were identified) and denote the dataset by $\acc{\prt{\ve{x}^{(i)}, \ell^{(i)}}, i = 1 \enum N}$ where $\ell^{(i)} = \acc{-1, \, 1}$ are the labels of the training points.

Given this training set, the support vector classifier (SVC) prediction for any yet-to-be observed sample reads \citep{Smola2004}
\begin{equation}\label{eq:02}
	\mathcal{M}^{\text{SVC}}\prt{\ve{x}} = \sum_{i=1}^N \alpha_i \, \ell^{(i)} \, k \prt{\ve{x}^{(i)} , \ve{x} ; \ve{\theta}} + b,
\end{equation}
where $\acc{\ve{\alpha},b}$ are parameters to calibrate. The coefficients $\alpha_i$, some of which are the so-called \emph{support vectors}, and the offset parameter $b$ are obtained by solving a quadratic optimization problem 
\begin{equation}\label{eq:03}
	\begin{split}
		\min_{\ve{\alpha}} \quad & \frac{1}{2} \ve{\alpha}^T \prt{ \widetilde{\ve{K}} \ve{Y} \ve{Y}^T } \ve{\alpha} + \ve{h}^T \ve{\alpha} \\
		\text{subject to:} \quad & \ve{\alpha}^T \ve{Y} = 0, \qquad \alpha_i \geq 0, \qquad i = \acc{1, \ldots, N},
	\end{split}
\end{equation}
where $\ve{h} = \acc{-1 \enum -1}$ is a column vector of size $N$ and $\widetilde{\ve{K}} = \ve{K} + 1/C \ve{I}_N$ with $C>0$ being a penalty term. The matrix $\ve{K}$ is the so-called Gram matrix built by evaluating the parameterized kernel function on pairs of points of the training data set, such that $K_{ij} = k\prt{\ve{x}^{(i)},\ve{x}^{(j)} ; \ve{\theta}}, i,j \in \acc{1 \enum N}$. Multiple kernels have been used in SVM. In this work, we consider the Gaussian kernel defined by
\begin{equation}
 k\prt{\ve{x}^{(i)},\ve{x}^{(j)} ; \ve{\theta}} = \prod_{l=1}^{M}\exp \bra{-\frac{1}{2} \prt{\frac{x^{(i)}_l - x^{(j)}_l}{\theta_l^2}}^2}.
\end{equation}

The hyperparameters of this model are the penalty term $C$ which controls the penalty incurred for misclassifying a training point and the kernel parameter $\ve{\theta}$ which controls, among others, the smoothness of the separating hyperplane. They are both estimated in this work by minimizing the span estimate of the leave-one-out error \citep{Vapnik2000,Chapelle2002} using the covariance-matrix adaptation evolution scheme (CMA-ES) (See \citet{Arnold2012,MoustaphaJRUES2018,UQdoc_14_111} for details). 

\subsubsection{Extension to multi-class classification}
Let us now consider the case when the classification task aims at categorizing data with a set of $K>2$ labels, where each label is defined as $\ell^{(i)} = \ell_k$ if the original training pair $\acc{\ve{x}^{(i)}, y^{(i)}}$ belongs to the cluster $\mathcal{C}_k$. 

The most popular approach to tackle this multi-class problem is to reduce it to a series of binary classification problems that can be solved using a standard SVM algorithm. The two most popular approaches are the \emph{one-against-all} and the \emph{one-vs-one} decomposition schemes \citep{Hastie1997,Moreira1998}. In the former, one binary problem is derived for each class $k$ by assigning one label, say the positive one, to all samples such that $\ell^{(i)} = \ell_k$ and the negative label to all the other samples. In the one-vs-one approach, binary classifiers considering all pairs of labels and ignoring all other samples are built. This leads to a total of $K(K-1)/2$ classifiers, which is larger than the $K$ classifiers required by the one-against-all approach. However, such classifiers are trained on a noticeably smaller subset of the training samples making the overall procedure computationally efficient despite the larger number of classifiers to build. 

Both approaches can be generalized, or somehow combined, using concepts of the \emph{error correcting output codes} (ECOC) \citep{Dietterich1995}.  The recombination of the binary classifiers into a final one can be achieved either by a simple voting system or by considering the posterior probabilities derived from each classifier. In this work, we consider the one-vs-one approach with a final voting system thanks to its simplicity and efficiency. We note that in case of equal voting between two classes, we heuristically choose the class that was predicted with the classifier that considered the two classes of interest. 

\subsubsection{Posterior probabilities}
As mentioned in Section~\ref{sec:setup}, the soft recombination of the final predictor requires some weights which are proportional to the probability that the sample belongs to a given class. In case of SVM, such weights can be derived by computing posterior probabilities derived from the classifier. In practice, this can be achieved by post-processing the output of the classifier using a sigmoid function as proposed by \citet{Platt2000}:
\begin{equation}\label{eq:p_i:bin}
	\mathbb{P}\prt{\ell\prt{\ve{x}} = 1 | \mathcal{M}^{\text{SVC}}\prt{\ve{x}}} = \frac{1}{1 + \exp\prt{A \, \mathcal{M}^{\text{SVC}}\prt{\ve{x}}  + B}},
\end{equation}
where the coefficients $A$ and $B$ are calibrated by solving a regularized maximum likelihood problem. In this work, we use an efficient numerical implementation proposed by \citet{Lin2007}.

There have been many attempts to extend these probabilities to multi-class problems \citep{Hastie1997,Moreira1998,Wu2004,Wang2008}. Let us denote by
\begin{equation}
	p_{ij} = \mathbb{P}\prt{\ve{x} \in \mathcal{C}_i |\, \ve{x} \in \mathcal{C}_i \cup \mathcal{C}_j}
\end{equation}
the posterior probability provided by the classifier that discriminates between the classes $\mathcal{C}_i$ (positive) and $\mathcal{C}_j$ (negative). Note however that we are interested in the overall probability of belonging to a class given all possible classes, \emph{i.e.} $p_i = \mathbb{P}\prt{\ve{x} \in \mathcal{C}_i}$. \citet{Moreira1998} proposed estimating this probability by combining the partial ones, \emph{i.e.},
\begin{equation}\label{eq:p_i:estim}
	\widehat{p}_i = \frac{2}{k(k-1)} \sum_{j \neq i, j = 1}^{K}p_{ij}
\end{equation}
This value is however flawed, as it accounts for spurious probabilities defined by classifiers discriminating two classes, none of which being the true one. 

Using Bayes theorem, it can however be noted that
\begin{equation}
	p_i = \mathbb{P}\prt{\ve{x} \in \mathcal{C}_i} = \mathbb{P}\prt{\ve{x} \in \mathcal{C}_i |\, \ve{x} \in \mathcal{C}_i \cup \mathcal{C}_j} \mathbb{P}\prt{\ve{x} \in \mathcal{C}_i \cup \mathcal{C}_j},
\end{equation}
which, by averaging over all combinations of $i$ and $j$, leads to the following system of equations:
\begin{equation}\label{eq:pisyst}
	p_i = \frac{1}{k-1} \sum_{j \neq i, j = 1}^{K} p_{ij} \prt{p_i+p_j},
\end{equation} 
since $\mathbb{P}\prt{\ve{x} \in \mathcal{C}_i \cup \mathcal{C}_j} = \prt{p_i+p_j}$. 
\citet{Wu2004} noted that this system of equations can be written in a matrix form 
\begin{equation}\label{eq:MarkovChain}
	\ve{p} = \ve{T} \ve{p},
\end{equation}
where $\ve{p} = \acc{p_1, \ldots, p_K}^T$ and $\ve{T}$ is a $K \times K$ matrix whose elements read
\begin{equation}
	\begin{split}
		T_{ij} = 
		\left\{ \begin{array}{ll}
		\frac{1}{k-1} \, p_{ij} &  \text{if} \:i \neq j,\\
			\frac{1}{k-1} \, \sum_{j \neq i, j = 1}^{K} p_{ij}  &  \text{if} \: i = j.\\
		\end{array} \right.
	\end{split}
\end{equation}
\citet{Wu2004} then noted that there exists a finite Markov chain whose transition matrix is $\ve{T}$, since $\sum_{j=1}^K T_{ij} = 1$ and $ 0 \leq T_{ij} \leq 1$. Further assuming that $p_{ij} > 0$ for any $i, \, j \in \acc{1, \ldots, K}$ implies that $T_{ij} > 0$, which ensures that the Markov chain is irreducible and aperiodic. In fine, these conditions guarantee that Eq.~\eqref{eq:MarkovChain} defines a Markov chain whose stationary distribution exists and is unique.

Taking advantage of the fact that $\ve{T}$ is a transition kernel and $\ve{p}$ is the stationary distribution of the corresponding Markov chain,  we cast Eq.~\eqref{eq:pisyst}  in an iterative scheme 
\begin{equation}\label{eq:p_i:iter}
	p_i^{(t+1)} = \frac{1}{k-1} \sum_{j \neq i, j = 1}^{K} p_{ij} \prt{p_i^{(t)}+p_j^{(t)}},
\end{equation}
where the initial values $p_i^{(0)}, \, i = \acc{1, \ldots K}$  using the estimate in Eq.~\eqref{eq:p_i:estim} and $p_{ij}$ are the partial probabilities obtained by the binary one-vs-one classifiers using Eq.~\eqref{eq:p_i:bin} . This chain eventually converges after a few iterations, generally with $t < 100$ in our examples, to the posterior probabilities estimates. 

\subsection{Regression using Kriging}\label{sec:KRG}
\subsubsection{Basics of Kriging}
The final ingredient considered in the proposed framework is Kriging a.k.a. Gaussian process model. It is used here to build local surrogates in the different regions identified by the clustering step. 

A Kriging model assumes that the model to approximate is of the form \citep{Santner2003,Rasmussen2006}
\begin{equation}\label{eq:05}
	\mathcal{M}\prt{\ve{x}} = \sum_{j=1}^p \beta_j f_j\prt{\ve{x}} + Z\prt{\ve{x}},
\end{equation}
where the first summand represents the \emph{trend} written here in a polynomial form using $p$ regressors $f_j$ with corresponding coefficients $\beta_j$. The second summand is a zero-mean stationary covariance process defined  by an auto-covariance function $\text{Cov}\bra{Z\prt{\ve{x}},Z\prt{\ve{x}^\prime}} = \sigma^2 R\prt{\ve{x},\ve{x}^\prime;\ve{\theta}}$ where $\sigma^2$ is the process variance and $R$ is an auto-correlation function parameterized by the vector $\ve{\theta}$. In this work, we consider an anisotropic Mat\'ern $5/2$ auto-correlation function defined by
\begin{equation}\label{eq:Matern-5_2}
R\prt{\ve{x}^{(i)},\ve{x}^{(j)} ; \ve{\theta}}  = \prod_{l=1}^{M} \bra{\prt{1 + \sqrt{5} \frac{\abs{x^{(i)}_l - x^{(j)}_l}}{\theta_l}
	+ \frac{5}{3} \prt{\frac{\abs{x^{(i)}_l - x^{(j)}_l}}{\theta_l}}^2}
	\exp \prt{-\sqrt{5}\frac{\abs{x^{(i)}_l - x^{(j)}_l}}{\theta_l}}}.
\end{equation}
The calibration of the model is performed by estimating the regression coefficients of the trend and the hyperparameters of the selected kernel that minimize a generalization error, herein using a maximum likelihood approach \citep{Santner2003,Bachoc2013b,Lataniotis2018}. 

Following this step, Kriging assumes that any unknown sample actually follows a normal distribution $\mathcal{N}\prt{\mu_{\widehat{\mathcal{M}}}, \sigma_{\widehat{\mathcal{M}}}^2 }$ where the mean is the actual prediction, while the standard deviation informs about the local accuracy of the prediction. The two quantities respectively read
\begin{equation}\label{eq:KRGpred}
	\begin{split}
		\mu_{\widehat{\mathcal{M}}}\prt{\ve{x}} & = \ve{f}^T\prt{\ve{x}} \widehat{\beta} + r\prt{\ve{x}} \ve{R}^{-1} \prt{\mathcal{Y} - \ve{F} \widehat{\beta}},\\
		\sigma_{\widehat{\mathcal{M}}}^2\prt{\ve{x}} & = \widehat{\sigma}^2 \prt{ 1 - \ve{r}\prt{\ve{x}}^T \ve{R}^{-1}  \ve{r}\prt{\ve{x}} + \ve{u}\prt{\ve{x}}^T \prt{\ve{F}^T \ve{R}^{-1} \ve{F}}^{-1} \ve{u}\prt{\ve{x}} },
	\end{split}
\end{equation}
where 
\begin{itemize}
	\item $\ve{u}\prt{\ve{x}} = \ve{F}^T \ve{R}^{-1} \ve{r}\prt{\ve{x}} - \ve{f}\prt{\ve{x}}$ has been introduced for convenience,
	\item $\widehat{\ve{\beta}} = \prt{\ve{F}^T \ve{R}^{-1} \ve{F}}^{-1} \ve{F}^T \ve{R}^{-1} \mathcal{Y}$ is the generalized least-square estimate of the regression coefficients $\ve{\beta}$,
	\item $\widehat{\sigma}^2 = \frac{1}{N} \prt{\mathcal{Y} - \ve{F} \widehat{\ve{\beta}}}^T \ve{R}^{-1} \prt{\mathcal{Y} - \ve{F} \widehat{\ve{\beta}}}$ is the estimate of the process variance,
	\item $\ve{F} = \acc{f_j\prt{\ve{x}^{(i)}}, \, j = 1 \enum p, \, i = 1 \enum n_0 }$ is the Vandermonde matrix, 
	\item $\ve{R}$ is the correlation matrix with $R_{ij} = R\prt{\ve{x}^{(i)},\ve{x}^{(j)};\ve{\theta}}$,
	\item $\ve{r}\prt{\ve{x}}$ is a vector gathering the correlation between the unknown sample $\ve{x}$ and the experimental design points and 
	\item $\mathcal{Y} = \acc{\mathcal{Y}^{(i)} = \mathcal{M}\prt{\ve{x}^{(i)}}, i = 1 \enum n_0}$ is the vector of available model responses.
\end{itemize}   

To account for the categorical variable, the \emph{compound symmetry} kernel defined by \citet{Pelamatti2020}
\begin{equation}
	R\prt{\ell^{(i)},\ell^{(j)}} = \left\{
	\begin{array}{ll}
		1 \quad \mbox{if} \quad \ell^{(i)} =\ell^{(j)},\\
		r \quad \mbox{if} \quad \ell^{(i)} \neq \ell^{(j)}, \\
	\end{array}
	\right.
\end{equation}
is considered. The parameter $r$ is computed here by embedding this kernel within a usual auto-correlation function for continuous variables with a tunable parameter $\theta_{\textrm{cat}}$ that can be calibrated in the same setting than the continuous parameters. More precisely, we consider a Gaussian kernel which then reads:
\begin{equation}\label{eq:Kcat}
	R\prt{\ell^{(i)},\ell^{(j)}; \theta_{\textrm{cat}}} = \exp\prt{-\frac{1}{2} \prt{\frac{S_{\ell^{(i)},\ell^{(j)}}}{\theta_{\textrm{cat}}}}^2},
\end{equation}
where $S_{\ell^{(i)},\ell^{(j)}} = 0$ if $\ell^{(i)} = \ell^{(j)}$ and $1$ otherwise. The final auto-correlation function is obtained by multiplying the $M+1$ one-dimensional auto-correlation functions \emph{i.e.},
\begin{equation}
		R\prt{\widetilde{\ve{x}}^{(i)},\widetilde{\ve{x}}^{{j}}, \widetilde{\ve{\theta}}} = \exp \prt{ -\frac{1}{2} \sum_{k=1}^{M} \prt{\frac{{\ve{x}}^{(i)} - {\ve{x}}^{(j)}}{\theta_k} }^2  -\frac{1}{2} \prt{\frac{S_{\ell^{(i)},\ell^{(j)}}}{\theta_{\textrm{cat}}}}^2 },
\end{equation}
where $\widetilde{\ve{\theta}} = \acc{\ve{\theta}, \, \theta_{\textrm{cat}}}$ and $\widetilde{\ve{x}}^{(i)} = \acc{\ve{x}^{(i)}, \, \ell^{(i)}}$.

\section{Examples}\label{sec:Examples}
The proposed algorithm is illustrated in this section with two analytical toy functions and an engineering problem related to a tensile membrane structure design. To assess its accuracy, we estimate the following two generalization errors using a validation set of size $N_\text{val} = 10^4$:
\begin{itemize}
	\item Normalized mean-square error:
	\begin{equation}\label{eq:12}
		NMSE = \frac{\sum_{i = 1}^{N_\text{val}} \prt{\mathcal{Y}_i - \widehat{\mathcal{Y}}_i}^2}{\sum_{i = 1}^{N_\text{val}} \prt{\mathcal{Y}_i - \bar{\mathcal{Y}}}^2}, 
	\end{equation}
	\item Mean absolute error:
	\begin{equation}\label{eq:12b}
		MAE =  \frac{1}{N_\text{val}}\sum_{i = 1}^{N_\text{val}} \abs{\mathcal{Y}_i - \widehat{\mathcal{Y}}_i}.
	\end{equation}
\end{itemize}
Furthermore, each analysis is repeated $20$ times in order to assess the robustness of the proposed algorithm with respect to the statistical uncertainty associated with the experimental designs.
\subsection{Manhattan function}
\begin{figure}[!ht]
	\centering
	\includegraphics[width=0.5\textwidth]{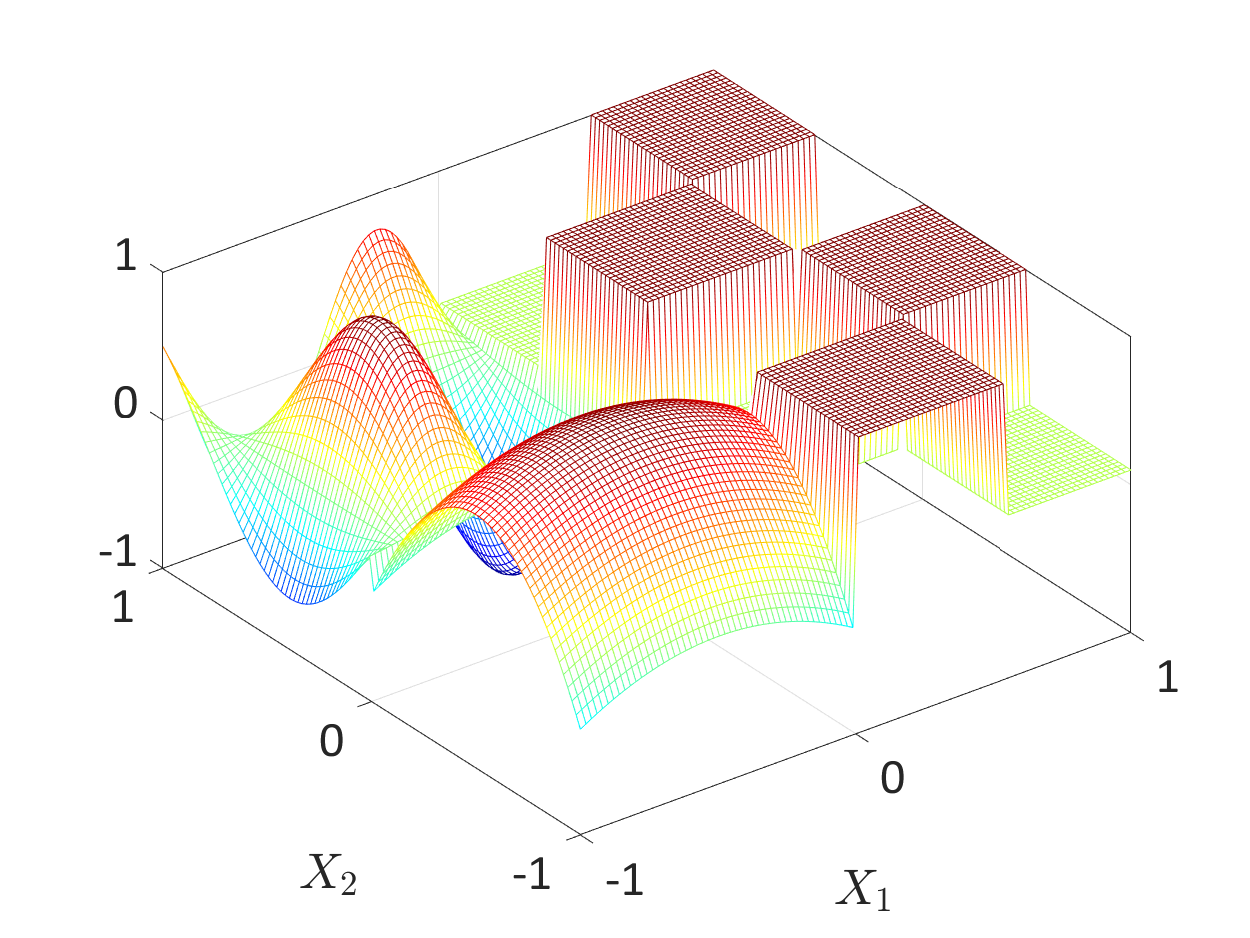}%
	\caption{Example 1 - Three-dimensional representation of the Manhattan function.}
	\label{fig:Manh3D}
\end{figure}

For this first validation example, we consider a two-dimensional function proposed by \citet{RaiThesis2015}. The function consists of three global regions, one of which is a checkerboard, and reads
\begin{equation*}
	\begin{split}
		\mathcal{M}\prt{\ve{x}} = 
		\left\{ \begin{array}{ll}
			\text{Checker board} \quad &  \text{if} \: x_1 \geq 0,\\
			\sin \prt{7 x_1} \cdot \sin \prt{4 x_2} ;\quad &  \text{if} \: x_1 \leq 0 \; \text{and} \; x_2 \leq 0,\\
			1 + \frac{2}{7} (2 x_1 + 1)^2 + (2  x_2 + 1)^2 ;\quad & \text{if} \: x_1 \leq 0 \; \text{and} \; x_2 \geq 0\\
		\end{array} \right.
	\end{split}
\end{equation*}
The checkerboard is made of smaller rectangular regions alternating the values of $0$ and $1$ as illustrated in Figure~\ref{fig:Manh3D}.

\begin{figure}[!ht]
	\centering
	\subfloat[Small - ED $\#1$]{\label{fig:Ex1:Clusters_a}\includegraphics[width=0.33\textwidth]{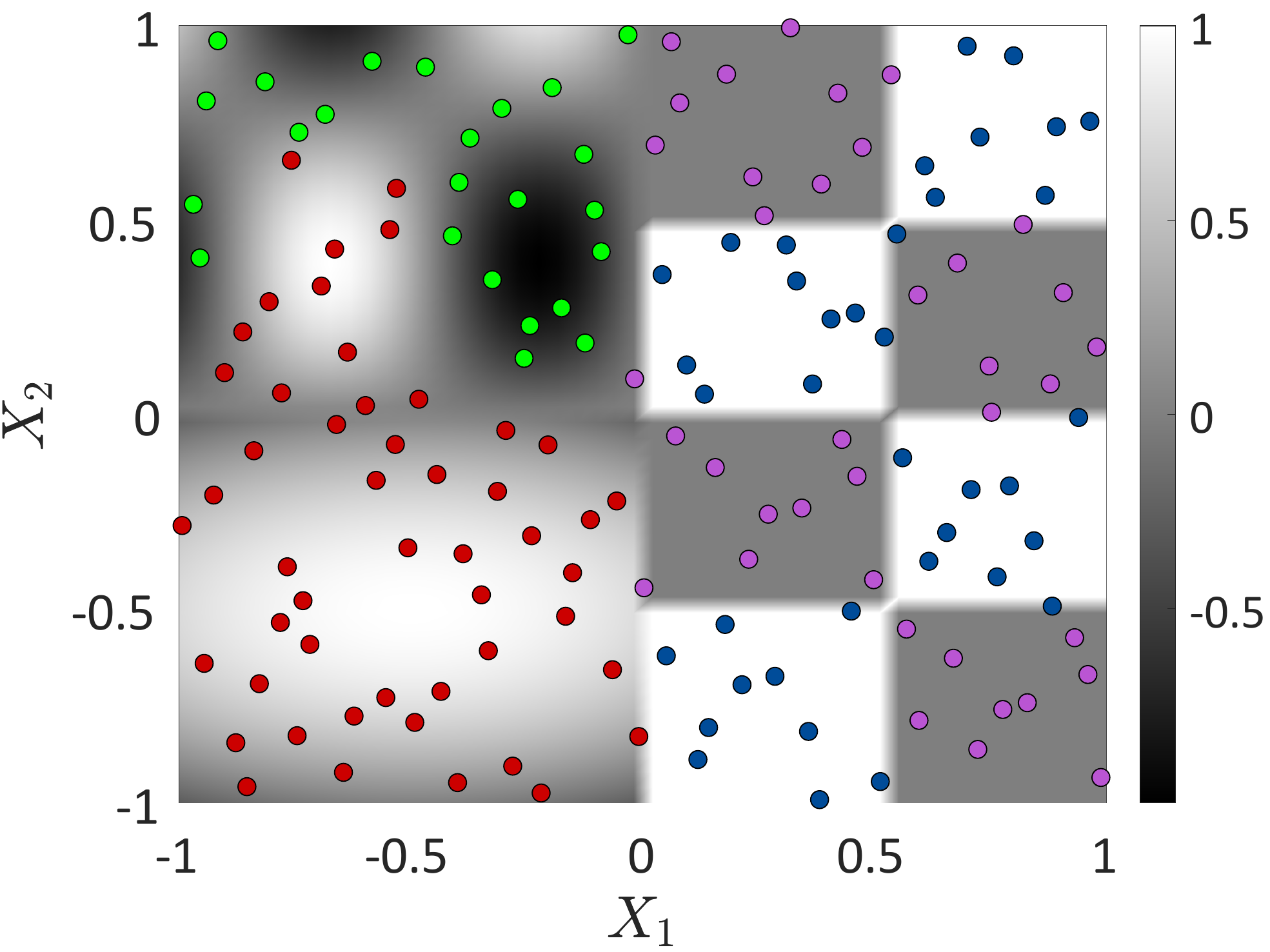}}%
	\subfloat[Medium - ED $\#1$]{\label{fig:Ex1:Clusters_b}\includegraphics[width=0.33\textwidth]{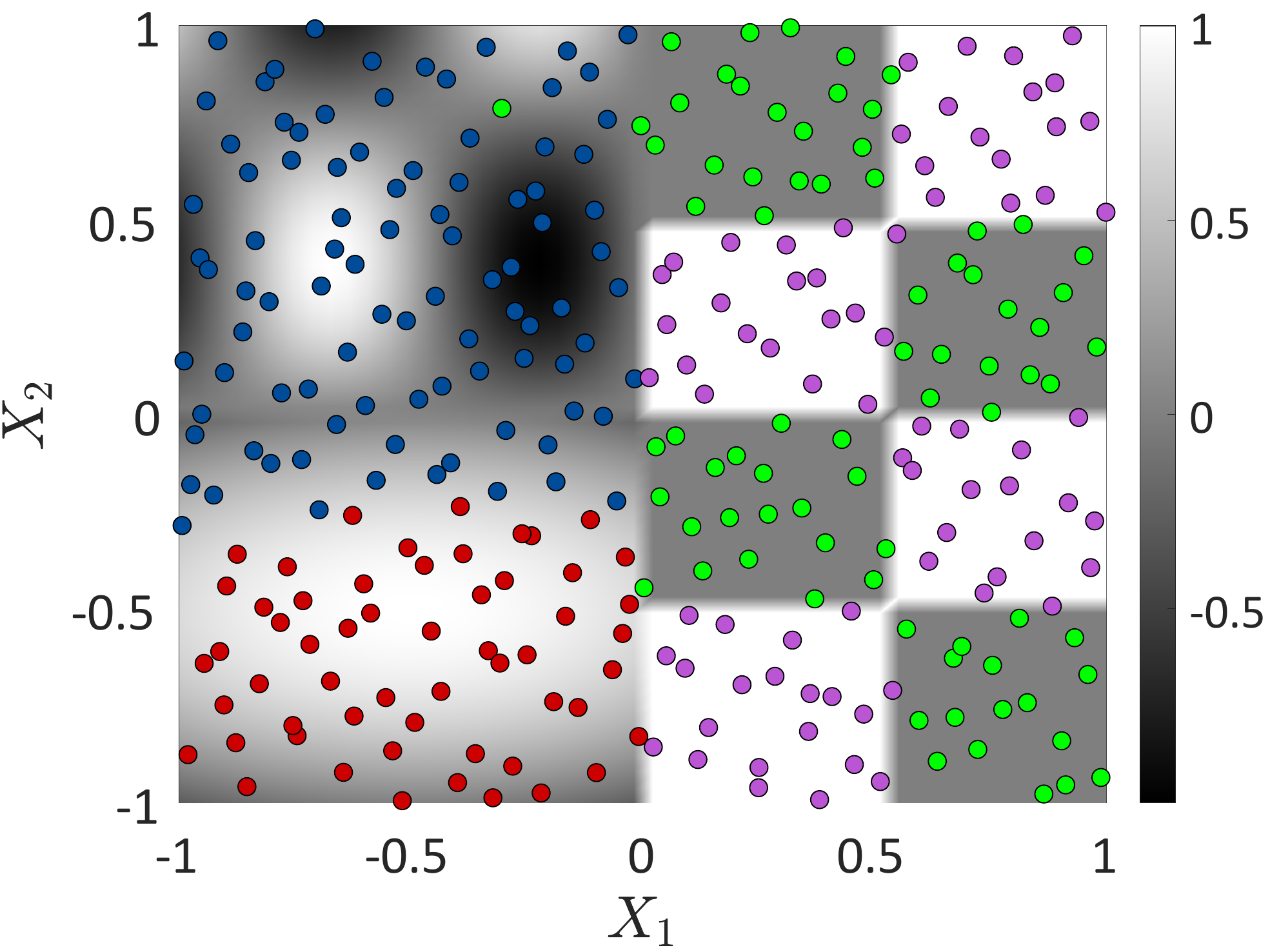}}%
	\subfloat[Large - ED $\#1$]{\label{fig:Ex1:Clusters_c}\includegraphics[width=0.33\textwidth]{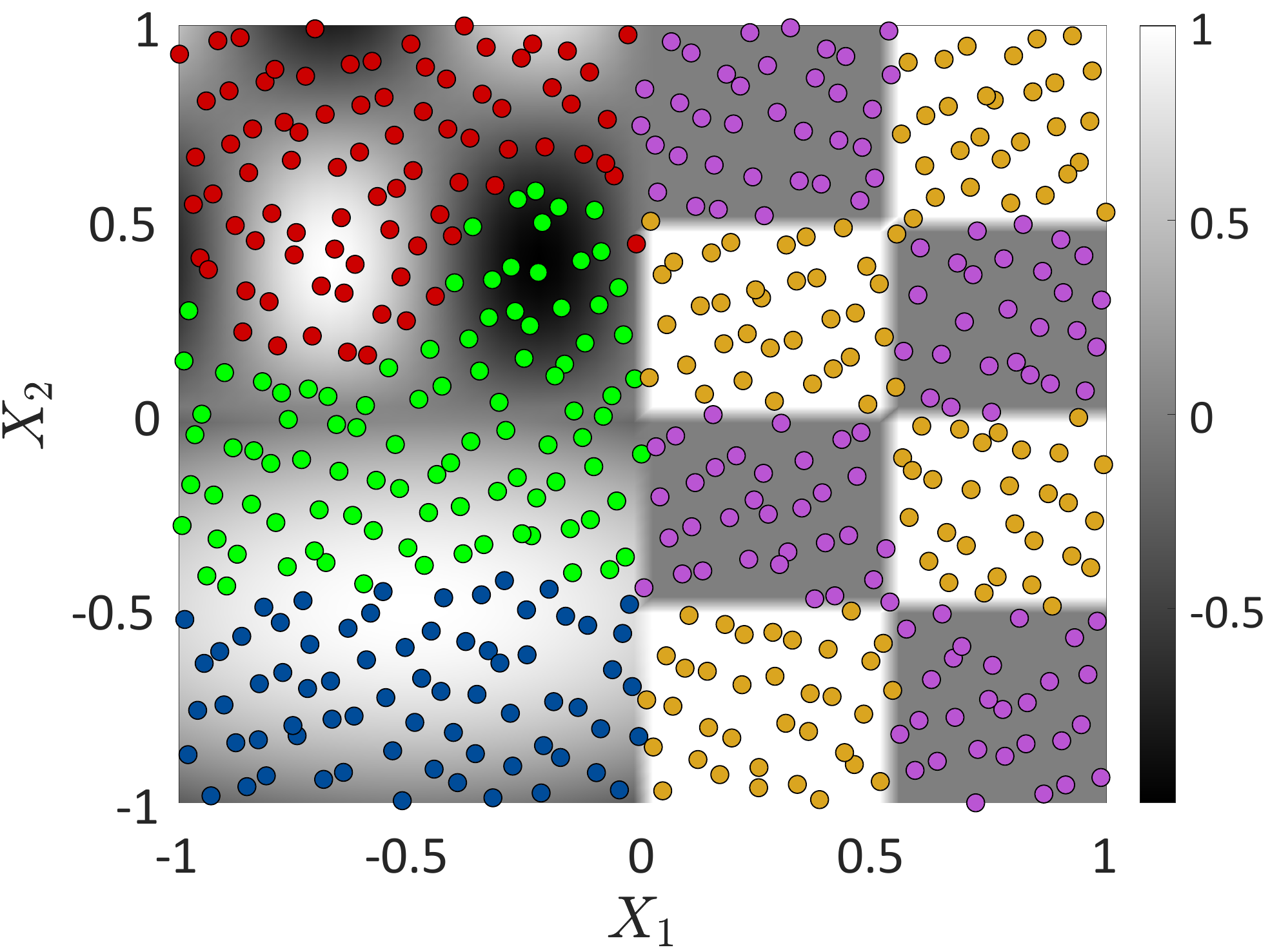}}%
	\\
	\subfloat[Small - ED $\#2$]{\label{fig:Ex1:Clusters_d}\includegraphics[width=0.33\textwidth]{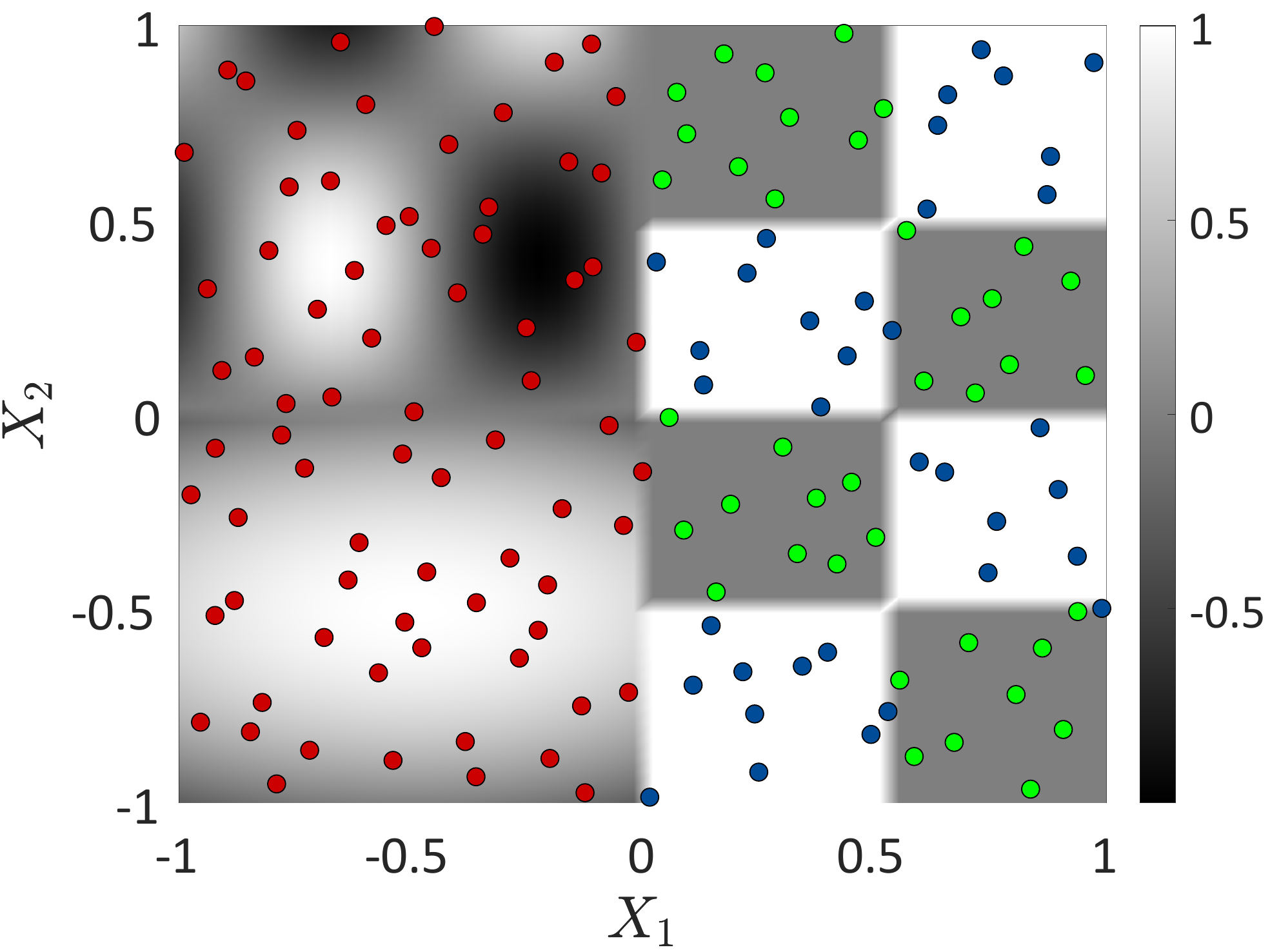}}%
	\subfloat[Medium - ED $\#2$]{\label{fig:Ex1:Clusters_e}\includegraphics[width=0.33\textwidth]{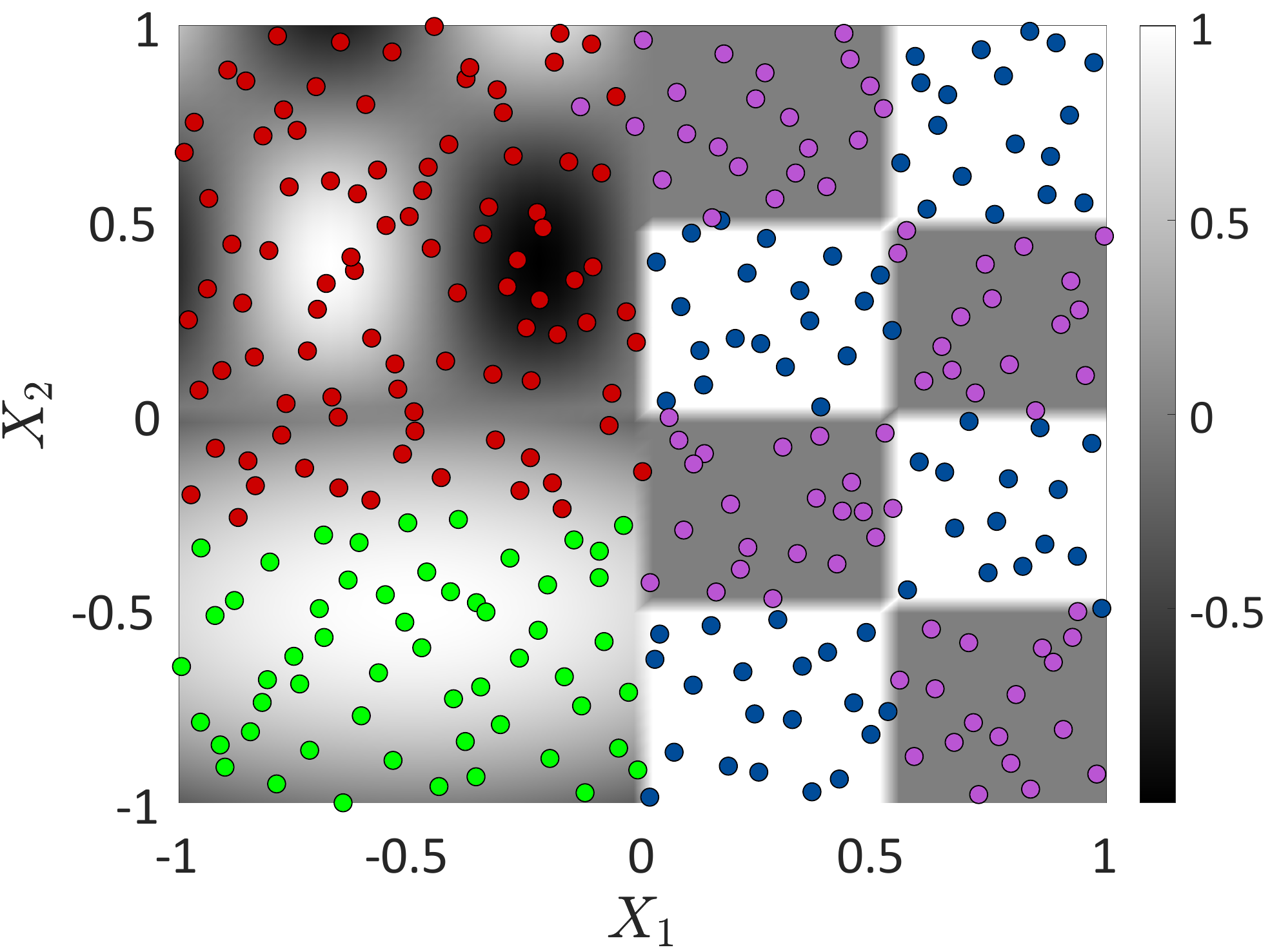}}%
	\subfloat[Large - ED $\#2$]{\label{fig:Ex1:Clusters_f}\includegraphics[width=0.33\textwidth]{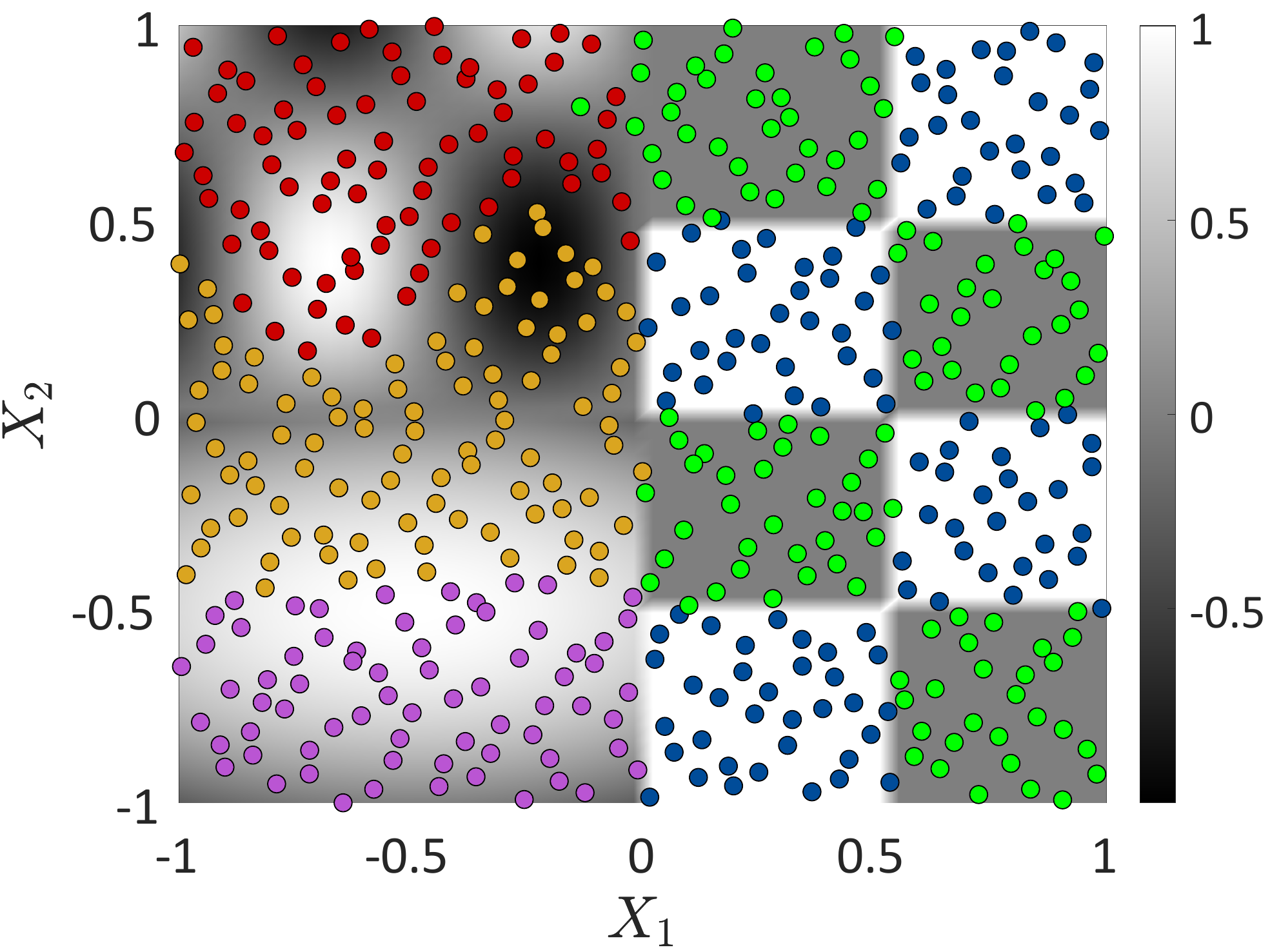}}%
	\caption{Example 1 - Clustering of the data by DPMM considering two repetitions of three experimental designs of increasing sizes.}
	\label{fig:Ex1:Clusters}
\end{figure}

In this section, we will illustrate each of the three steps of the proposed algorithm. We first start by showing how the clustering algorithm splits the data. Figure~\ref{fig:Ex1:Clusters} shows the clusters identified using three experimental designs of different sizes. The original model is built assuming $10$ regions where each of the squares in the checkerboard is considered as one region on its own. However, regardless of the experimental design, the clustering algorithm reduces the checkerboard into two regions, one with $y = 1$ and the other with $y=0$. This results in disconnected subdomains but as we will see in the next paragraph, this does not affect the overall prediction capability of the algorithm. Another important observation from the partitions in Figure~\ref{fig:Ex1:Clusters} is that the more data points, the more clusters are identified. For small datasets, the partition is quite sensitive to the data. However, the partition becomes more stable and robust as the data size is increased. 

\begin{figure}[!ht]
	\centering
	\subfloat[$\mathcal{C}_1$ vs. $\mathcal{C}_2 $]{\label{fig:Ex1:Classification_ovo_a}\includegraphics[width=0.33\textwidth]{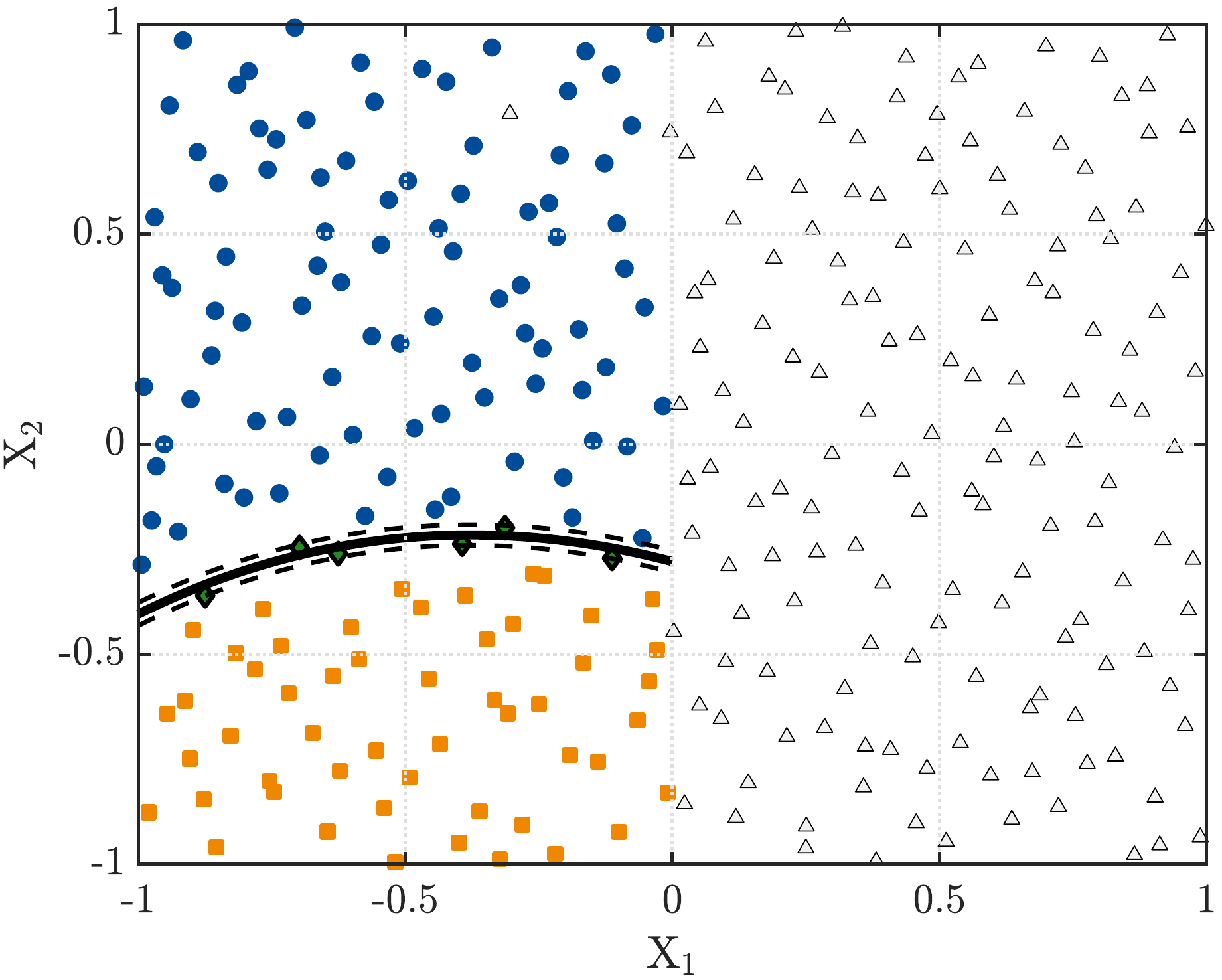}}%
	\subfloat[$\mathcal{C}_1$ vs. $\mathcal{C}_3 $]{\label{fig:Ex1:Classification_ovo_b}\includegraphics[width=0.33\textwidth]{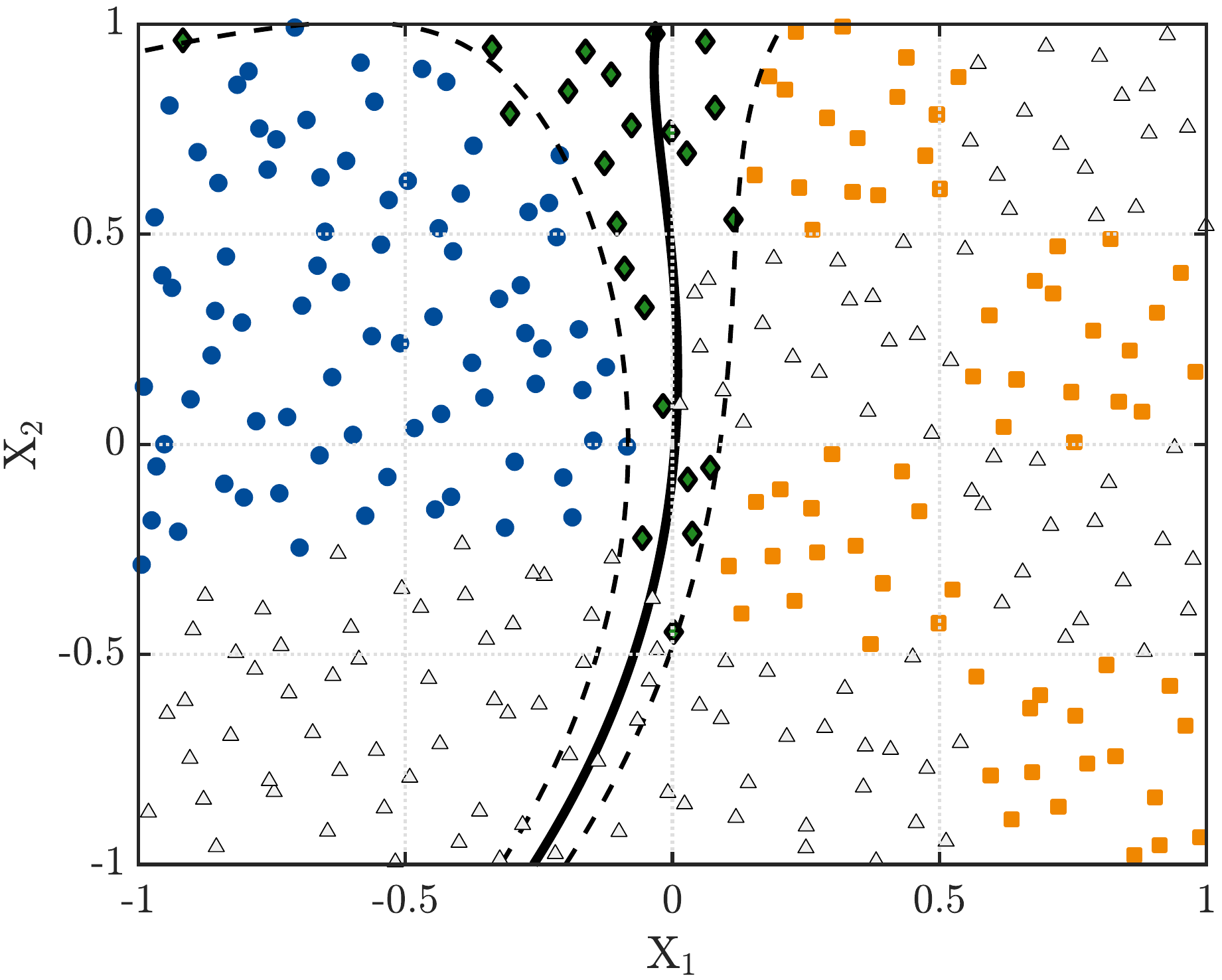}}%
	\subfloat[$\mathcal{C}_1$ vs. $\mathcal{C}_4 $]{\label{fig:Ex1:Classification_ovo_c}\includegraphics[width=0.33\textwidth]{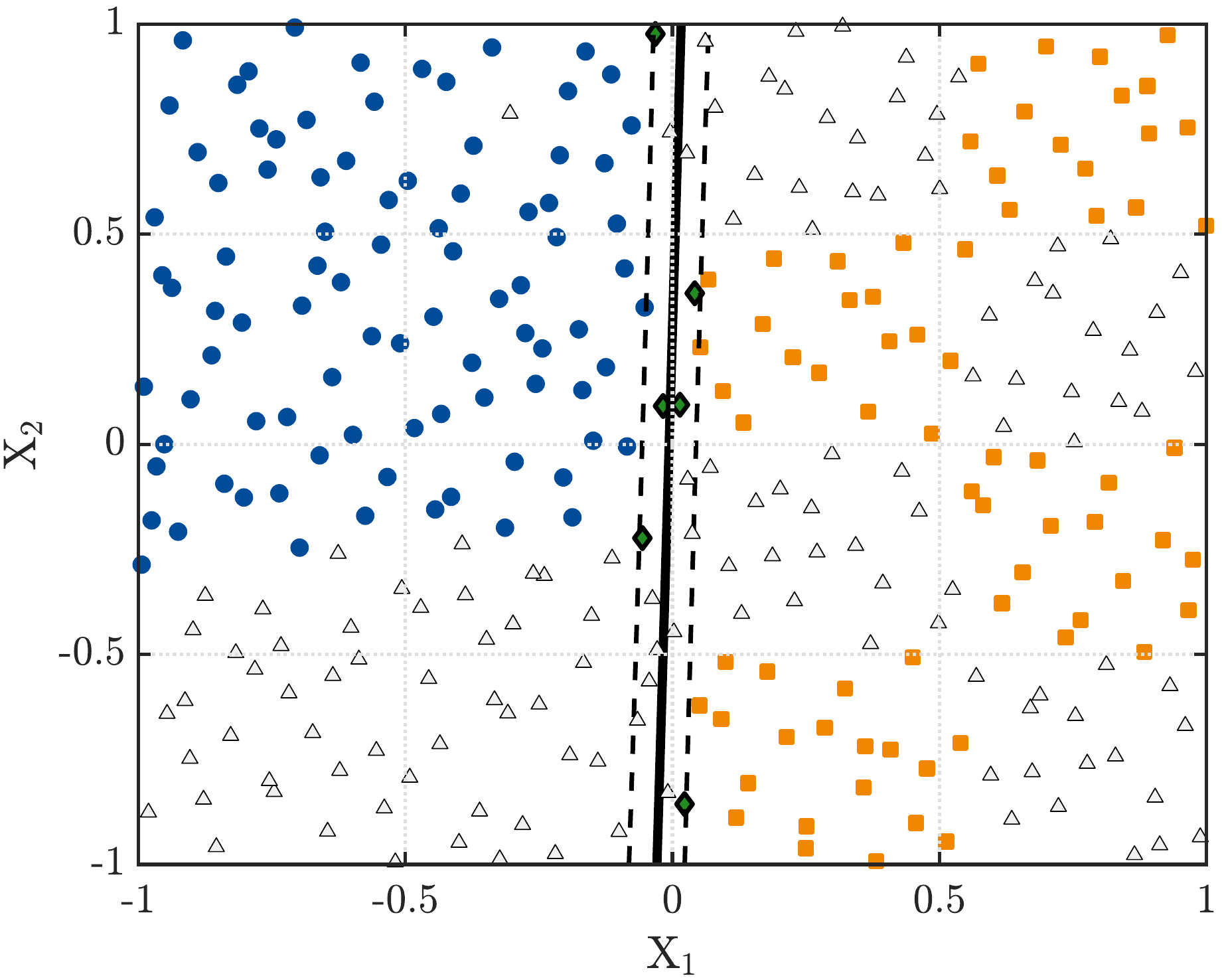}}%
	\\
	\subfloat[$\mathcal{C}_2$ vs. $\mathcal{C}_3 $]{\label{fig:Ex1:Classification_ovo_d}\includegraphics[width=0.33\textwidth]{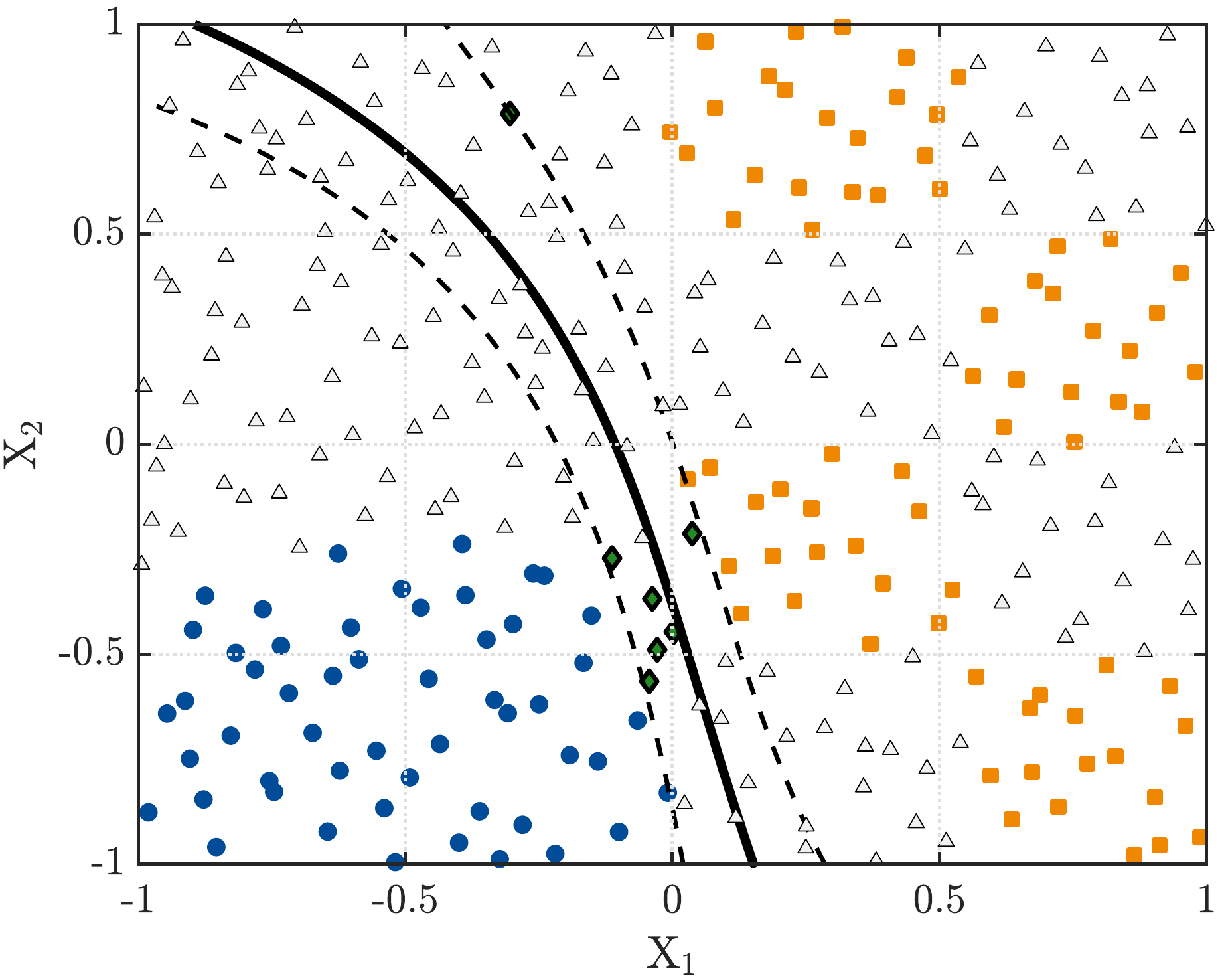}}%
	\subfloat[$\mathcal{C}_2$ vs. $\mathcal{C}_4 $]{\label{fig:Ex1:Classification_ovo_e}\includegraphics[width=0.33\textwidth]{Example1_Classification_onevone_3}}%
	\subfloat[$\mathcal{C}_3$ vs. $\mathcal{C}_4 $]{\label{fig:Ex1:Classification_ovo_f}\includegraphics[width=0.33\textwidth]{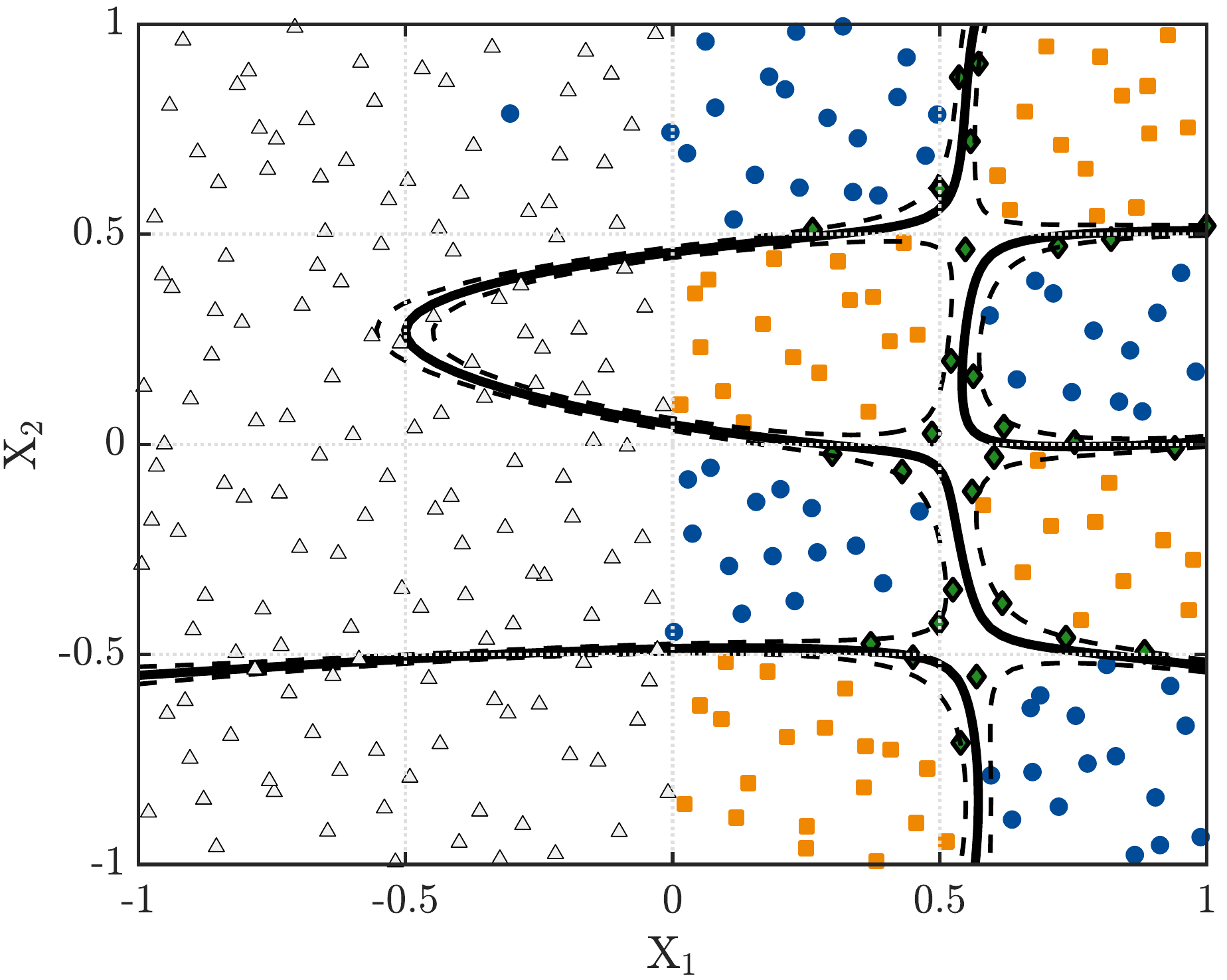}}%
	\caption{Example 1 - Pairwise classification of the data (with 4 clusters identified in Step 1 for the medium-size experimental design).}
	\label{fig:Ex1:Classification_ovo}
\end{figure}

Once the clusters are identified ($4$ different ones in the case of medium-size experimental design, and in the sequel), the inputs are labelled accordingly and binary classification is performed on each pair of classes. Figure~\ref{fig:Ex1:Classification_ovo} shows the resulting classifiers for one realization of the experimental design. The blue and orange points correspond to the positive and negative labels respectively, while the support vectors are highlighted in green. The thick line is the classifier, whereas the dashed ones delimit the margin. Finally, the gray triangles represent the data points that were ignored by the illustrated classifier. As expected, support vector machines are appropriately calibrated for the problem at hand. However, the choice of the Gaussian kernel may not be appropriate for the classification of $\mathcal{C}_3$ against $\mathcal{C}_4$ (Figure~\ref{fig:Ex1:Classification_ovo_f}) as it produces smooth boundaries whereas the original boundary results from a checkerboard with sharp edges. This does not substantially affect the results. However, better prediction could have been obtained by including the choice of the kernel in the model selection.

\begin{figure}[!ht]
	\centering
	\includegraphics[width=0.5\textwidth]{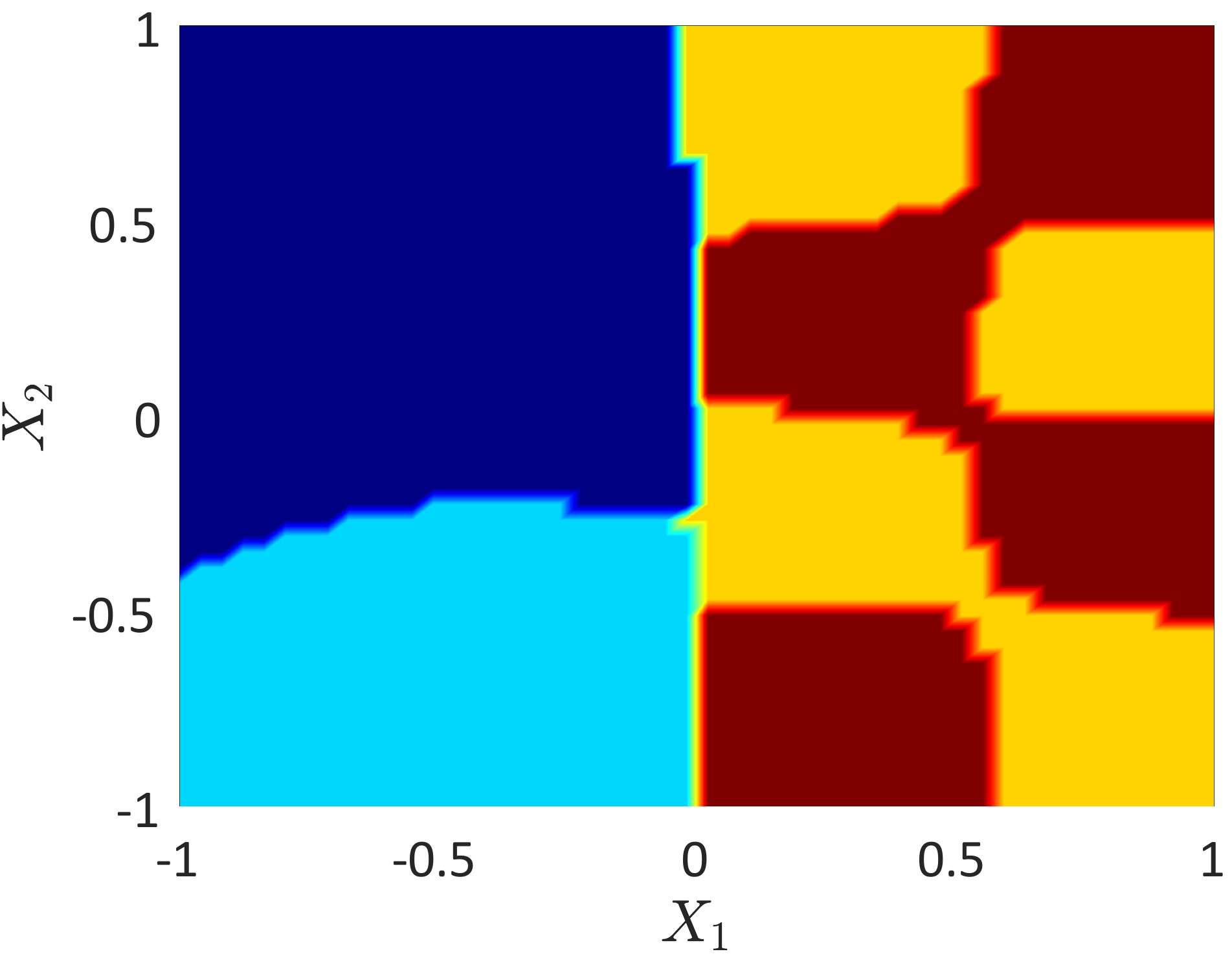}%
	\caption{Example 1 - Partition of the space in the $4$ regions using hard reconstruction.}
	\label{fig:Ex1:Class:Hard}
\end{figure}

\begin{figure}[!ht]
	\centering
	\subfloat[$\text{Prob}\bra{Y \in \mathcal{C}_1 }$]{\label{fig:Ex1:Class:Soft_a}\includegraphics[width=0.45\textwidth]{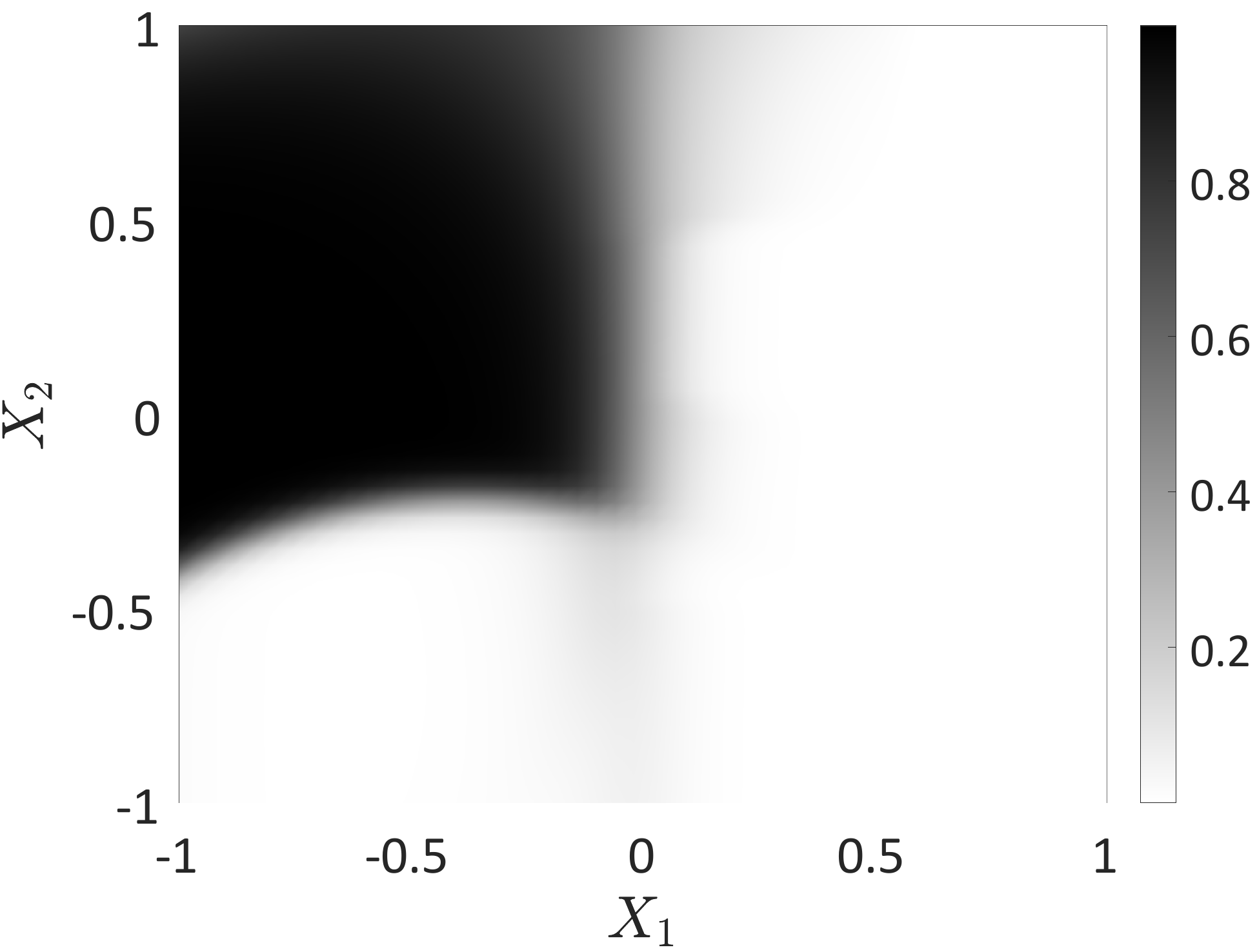}}%
	\subfloat[$\text{Prob}\bra{Y \in \mathcal{C}_2 }$]{\label{fig:Ex1:Class:Softn_b}\includegraphics[width=0.45\textwidth]{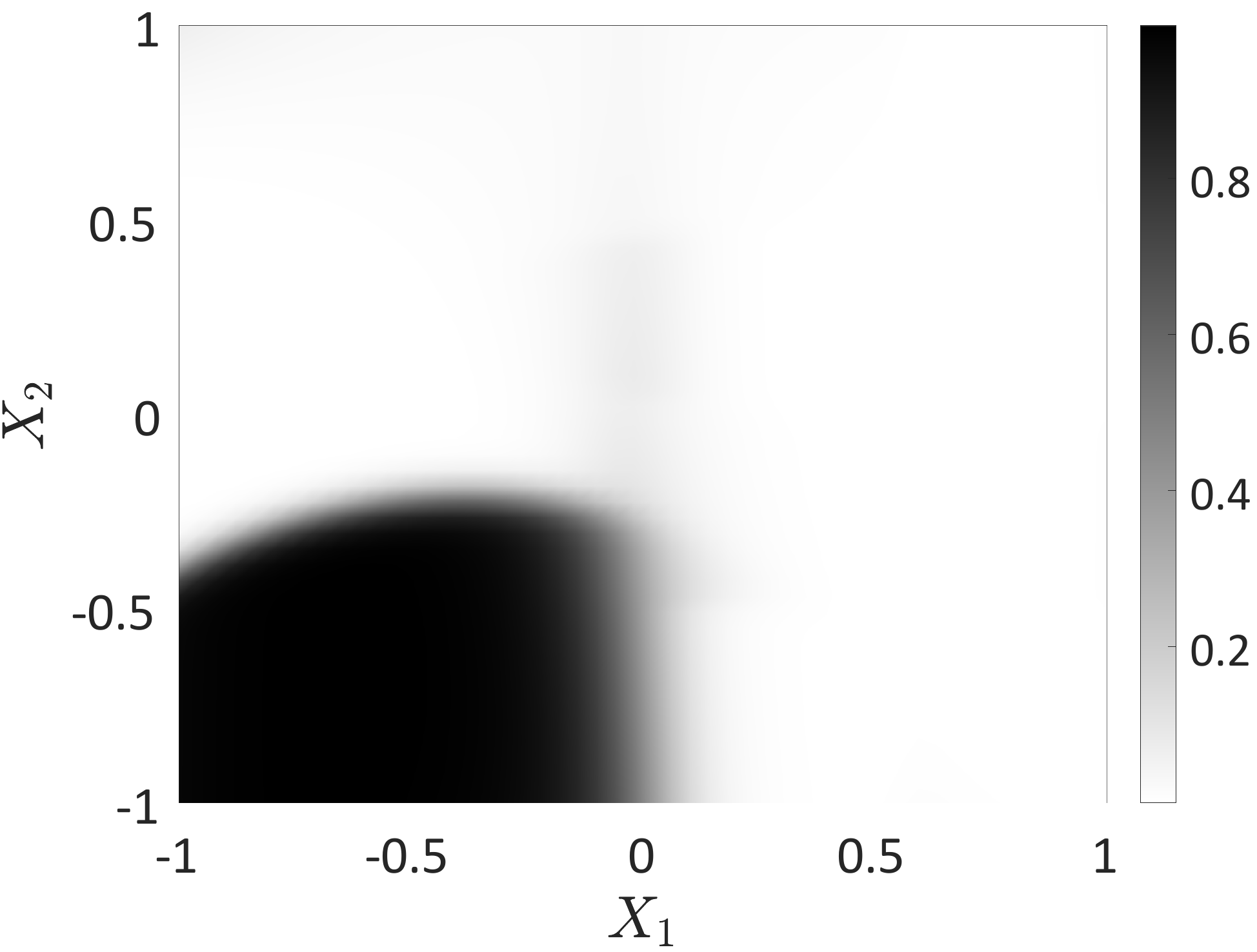}}%
	\\
	\subfloat[$\text{Prob}\bra{Y \in \mathcal{C}_3 }$]{\label{fig:Ex1:Class:Soft_c}\includegraphics[width=0.45\textwidth]{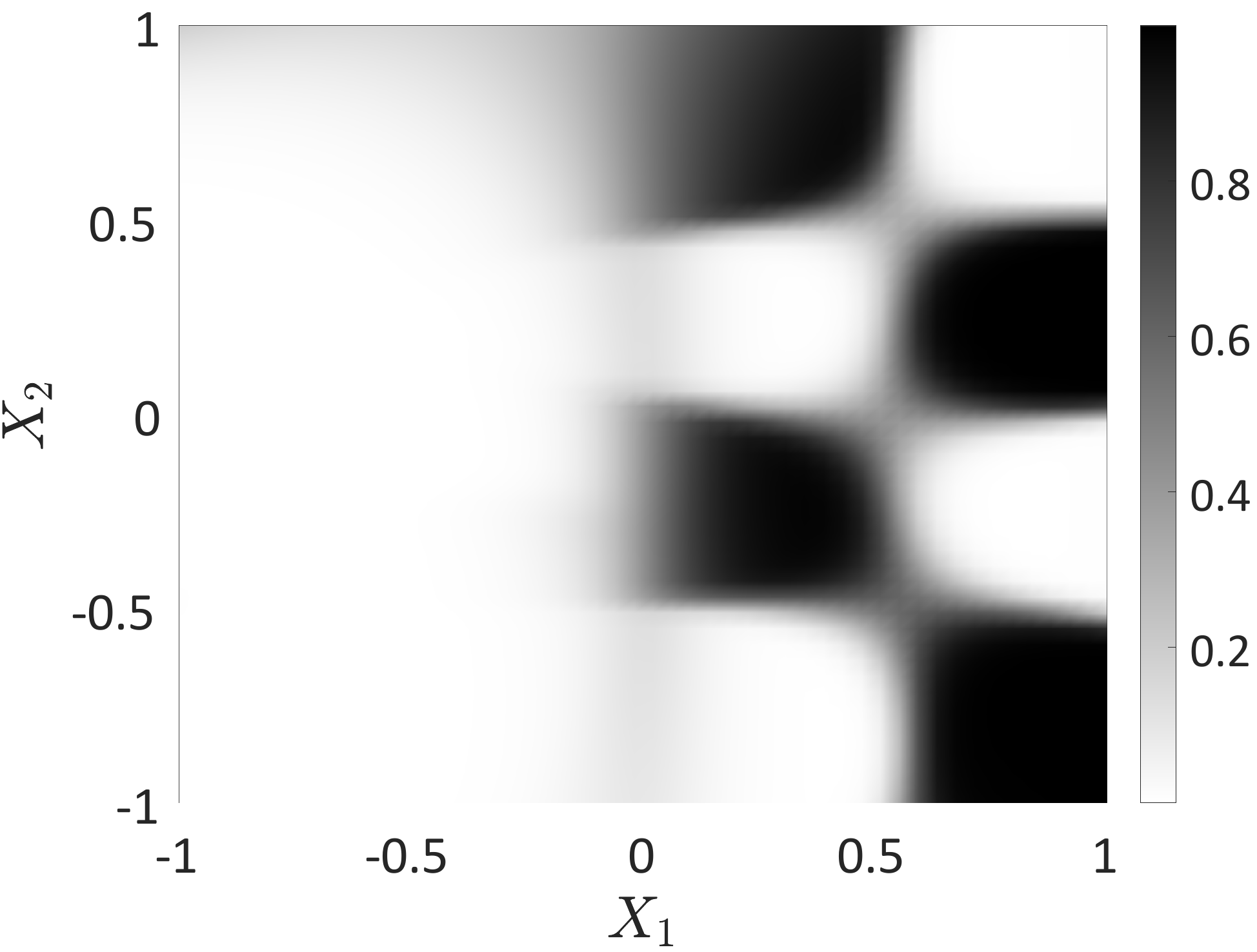}}%
	\subfloat[$\text{Prob}\bra{Y \in \mathcal{C}_4 }$]{\label{fig:Ex1:Class:Soft_d}\includegraphics[width=0.45\textwidth]{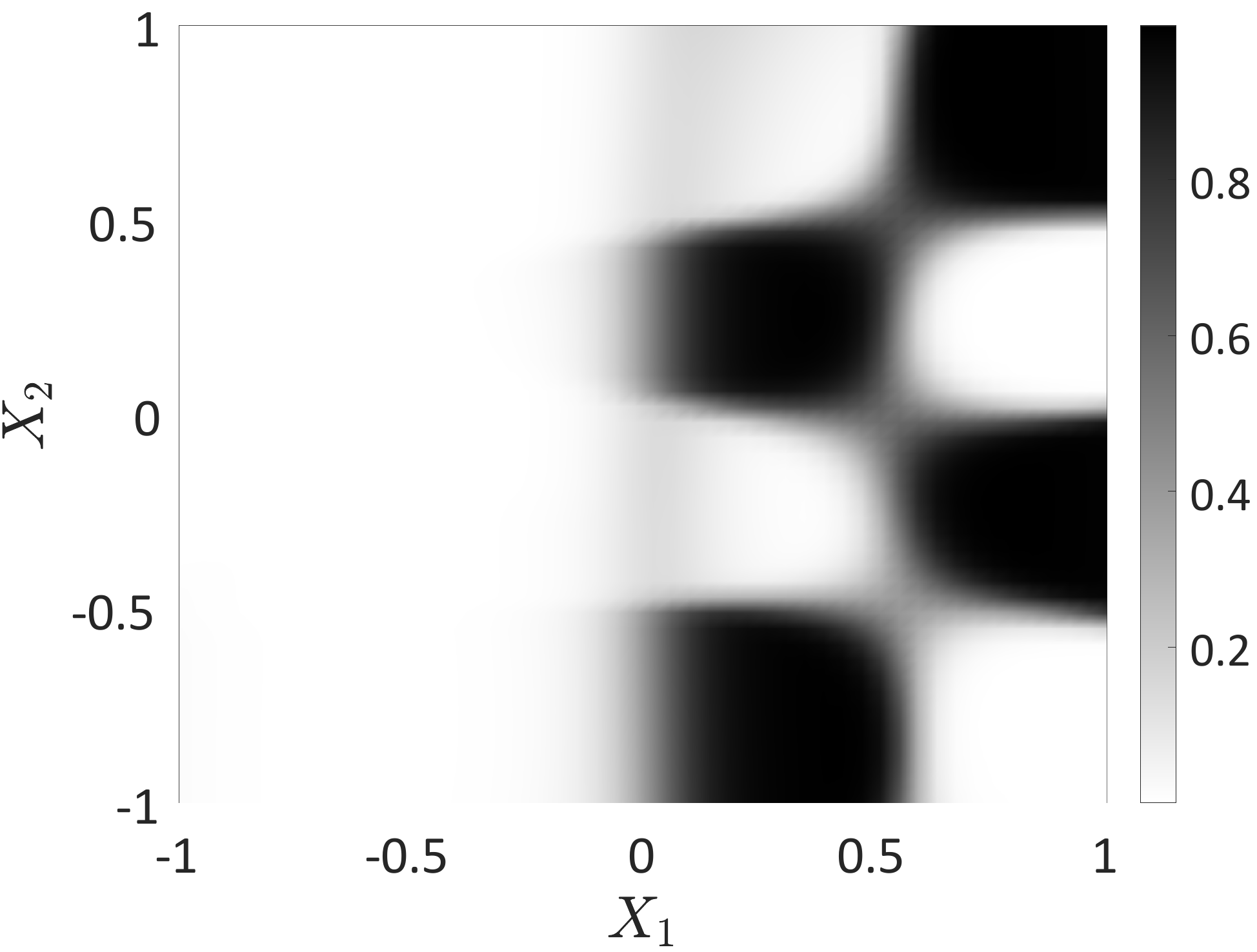}}%
	\caption{Example 1 - Partition of the space in the $4$ regions using soft    reconstruction.}
	\label{fig:Ex1:Class:Soft}
\end{figure}

The next step is then to recombine those predictions into a final one. In the hard reconstruction, a vote is carried out and the class that wins is the final prediction. The resulting partition of the input space is shown in Figure~\ref{fig:Ex1:Class:Hard}. Figure~\ref{fig:Ex1:Class:Soft} shows the soft reconstruction approach where each tile represents the probabilities of a given point to belong to a given class. The resulting classification is in accordance with the regions defined by the original model except for the boundaries of the checkerboard which present some slight deviations. Also, the boundary between the two regions where $\mathcal{M}$ is smooth (\ie, polynomial or sines) is not exactly the line $\acc{x_1 \leq 0, x_2 = 0}$.

\begin{figure}[!ht]
	\centering
	\subfloat[$NMSE$]{\label{fig:Ex1:Results_a}\includegraphics[width=0.45\textwidth]{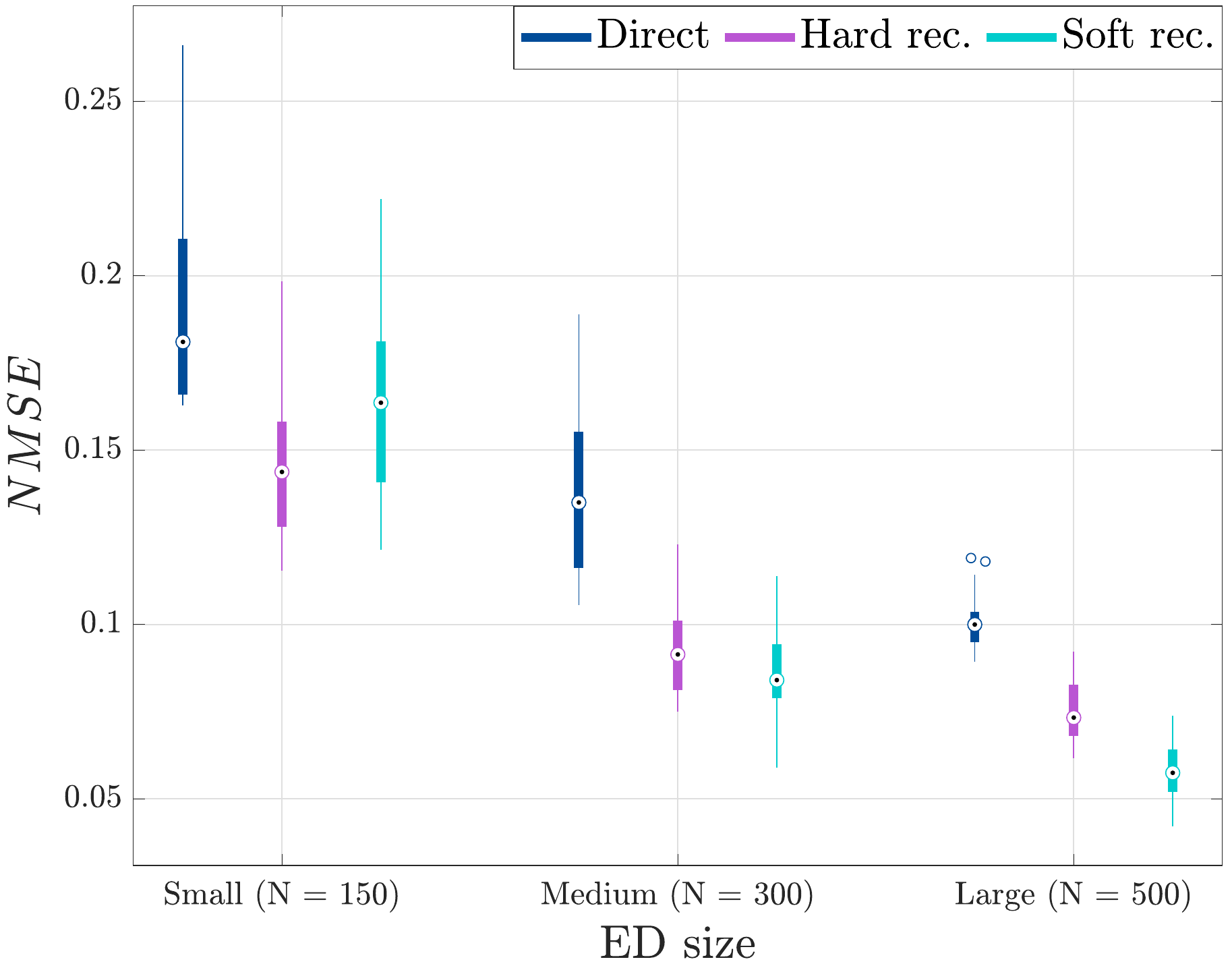}}%
	\subfloat[$MAE$]{\label{fig:Ex1:Results_b}\includegraphics[width=0.45\textwidth]{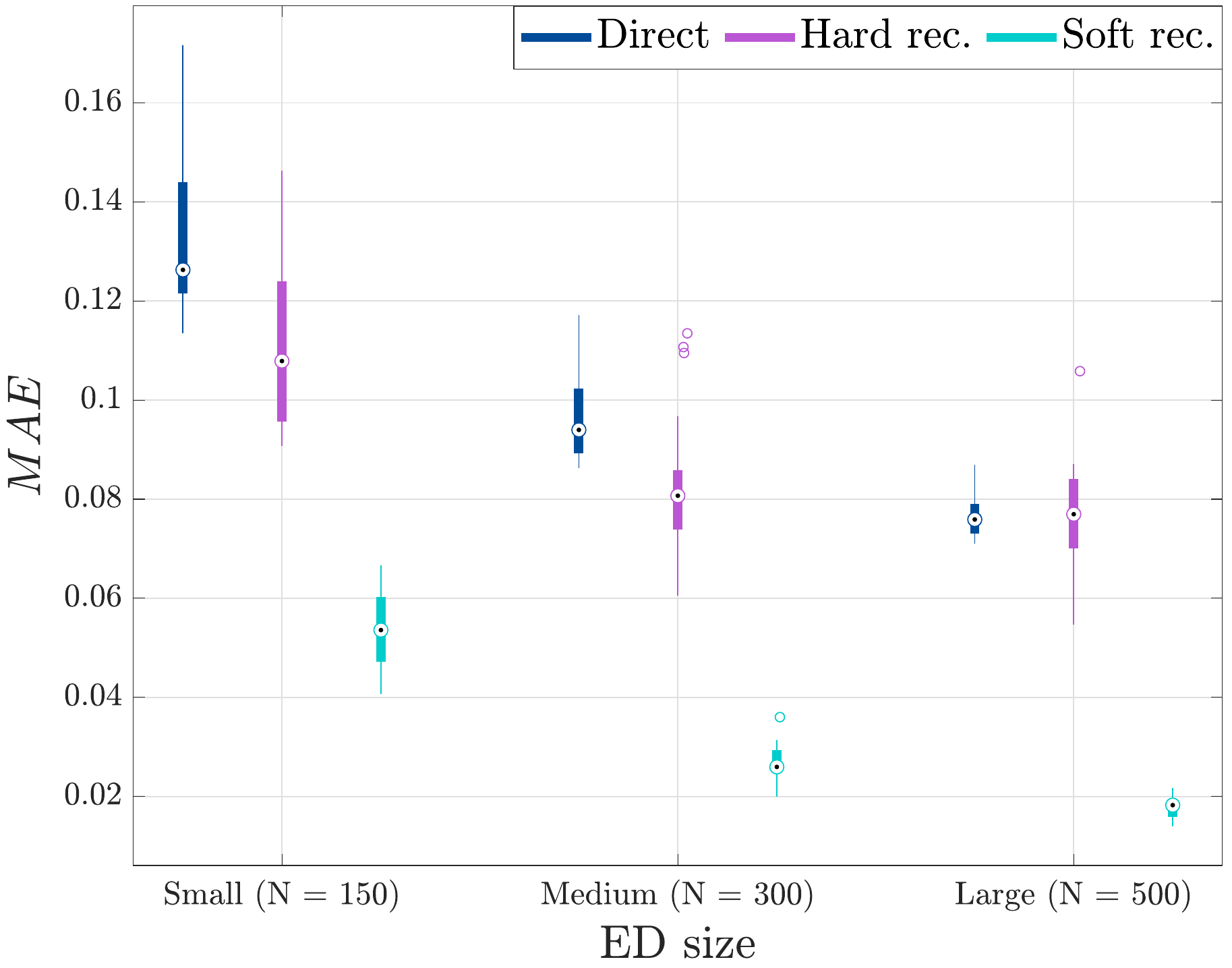}}%
	\caption{Example 1: Boxplots of the computed errors for various methods and experimental design sizes.}
	\label{fig:Ex1:Results}
\end{figure}

This partition of the input space is eventually used to build local Kriging surrogates to provide the final prediction. For this example, we repeat the analysis $20$ times where each repetition starts with a randomly sampled Sobol' sequence. Figure~\ref{fig:Ex1:Results} shows boxplots of the resulting errors for increasing sizes of the experimental design. For any ED size, both recombination techniques yield improved $NMSE$ and $MAE$. In general, the soft reconstruction also yields better prediction than the hard one. This is even more clear when considering the $MAE$ error. For this example, the prediction with categorical Kriging is not included, since it does not lead to good results. This is due to the fact that each region is fundamentally different from the other, hence using a single Kriging model, even with categorical variables, is not appropriate.

\clearpage

\subsection{Snap-through instability problem}
\begin{figure}[!ht]
	\begin{center}
		\includegraphics[width=0.4\textwidth]{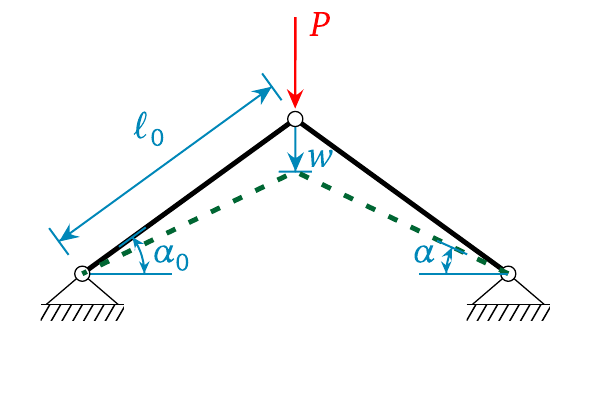}
		\caption{Illustration of the two-bar truss structure subject to snap-through.}
		\label{fig:truss}
	\end{center}
\end{figure}

This example is a mechanical problem related to the snap-through instability of a two-bar truss structure. The structure is loaded at its tip and responds linearly with small displacements until a critical point is reached. Past that point, the structure suddenly snaps through a new equilibrium point and resumes its small displacements. In this example, we consider as quantity of interest the displacement $w$ of the tip of the structure as illustrated in Figure~\ref{fig:truss} .

The load at the deformed position can be expressed as a function of the inclination angles at the initial position and deformed one, respectively denoted by $\alpha_0$ and $\alpha$, the bars cross-sectional areas $A$ and their constitutive material Youngs's modulus $E$
\begin{equation}\label{eq:13}
	P = - 2 E A \tan \prt{\alpha} \prt{\cos\prt{\alpha_0} - \cos\prt{\alpha}}.
\end{equation}
The corresponding displacement of the tip of the truss can then be computed as follows:
\begin{equation}\label{eq:14}
	w = l_0 \cos\prt{\alpha_0} \prt{ \tan\prt{\alpha_0} - \tan\prt{\alpha}}.
\end{equation}
In this example, we assume that the length of the bar $l_0 = 5$ m and the initial inclination angle $\alpha_0 = 10^\circ$ are deterministic. In contrast, the load, the Young's modulus and the cross section areas are assumed random and characterized by the distributions shown in Table~\ref{tab:example2}.
\begin{table}[!ht]
	\begin{center}
		\begin{tabular}{*{4}{c}}
			\hline
			Parameter & Distribution & Mean & C.o.V. \\
			\hline
			Load ($P$ in N) & Gumbel & $430$ & $0.20$ \\
			Young's modulus ($E$ in GPa)  & Lognormal & $210$ & $0.10$ \\
			Cross sectional area ($A$ in cm$^2$)  & Gaussian & $10$ & $0.05$ \\
			\hline
		\end{tabular}
	\end{center} 
	\caption{Truss snap-through problem: probabilistic input model.}
	\label{tab:example2}
\end{table}

\begin{figure}[!ht]
	\centering
	\subfloat[$NMSE$]{\label{fig:Ex2:Results_a}\includegraphics[width=0.45\textwidth]{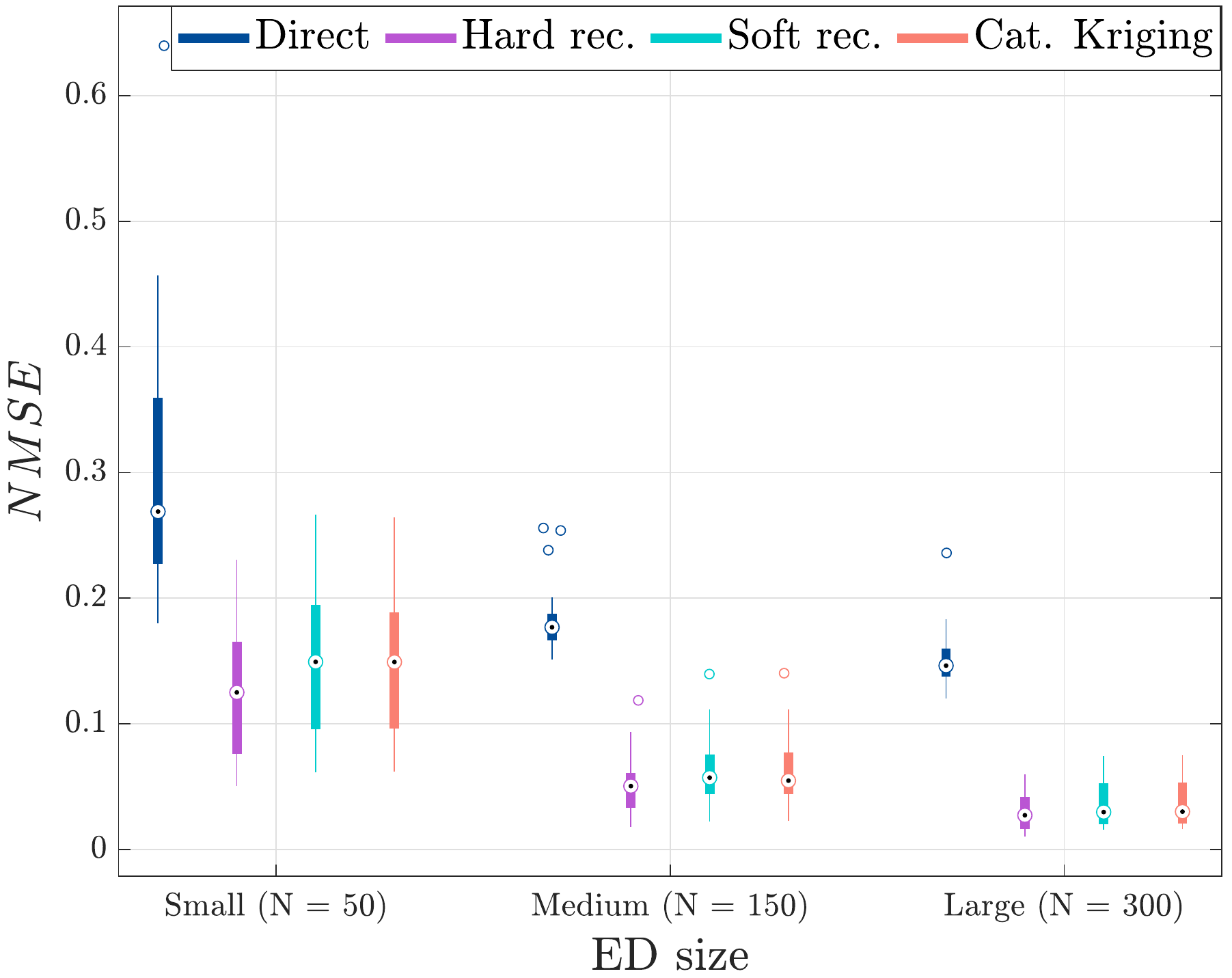}}%
	\subfloat[$MAE$]{\label{fig:Ex2:Results_b}\includegraphics[width=0.45\textwidth]{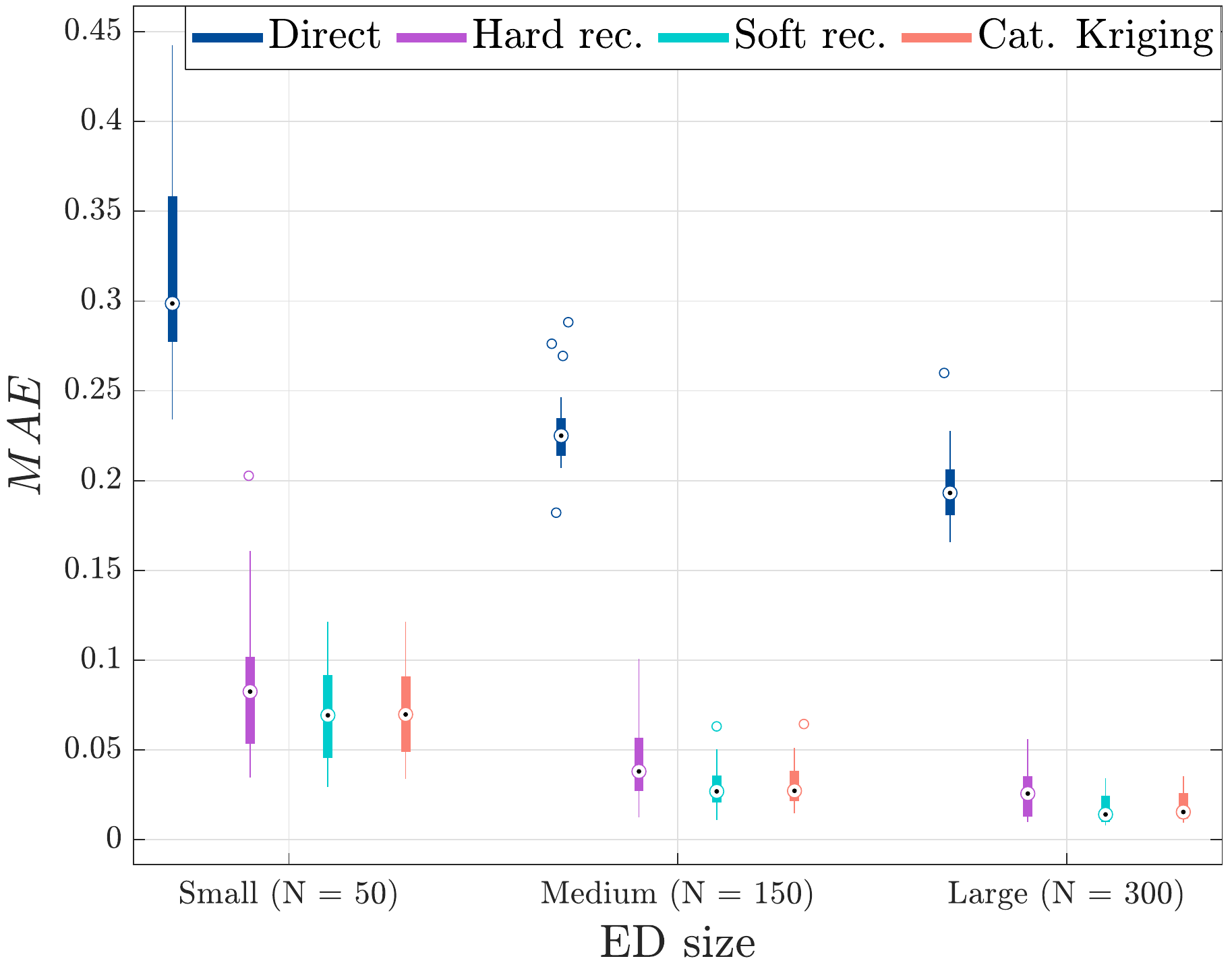}}%
	\caption{Example 2: Boxplots of the computed errors for various methods and experimental design sizes.}
	\label{fig:Ex2:Results}
\end{figure}

\begin{figure}[!ht]
	\centering
	\subfloat[Direct Kriging and hard reconstruction]{\label{fig:Ex2:YY_a}\includegraphics[width=0.45\textwidth]{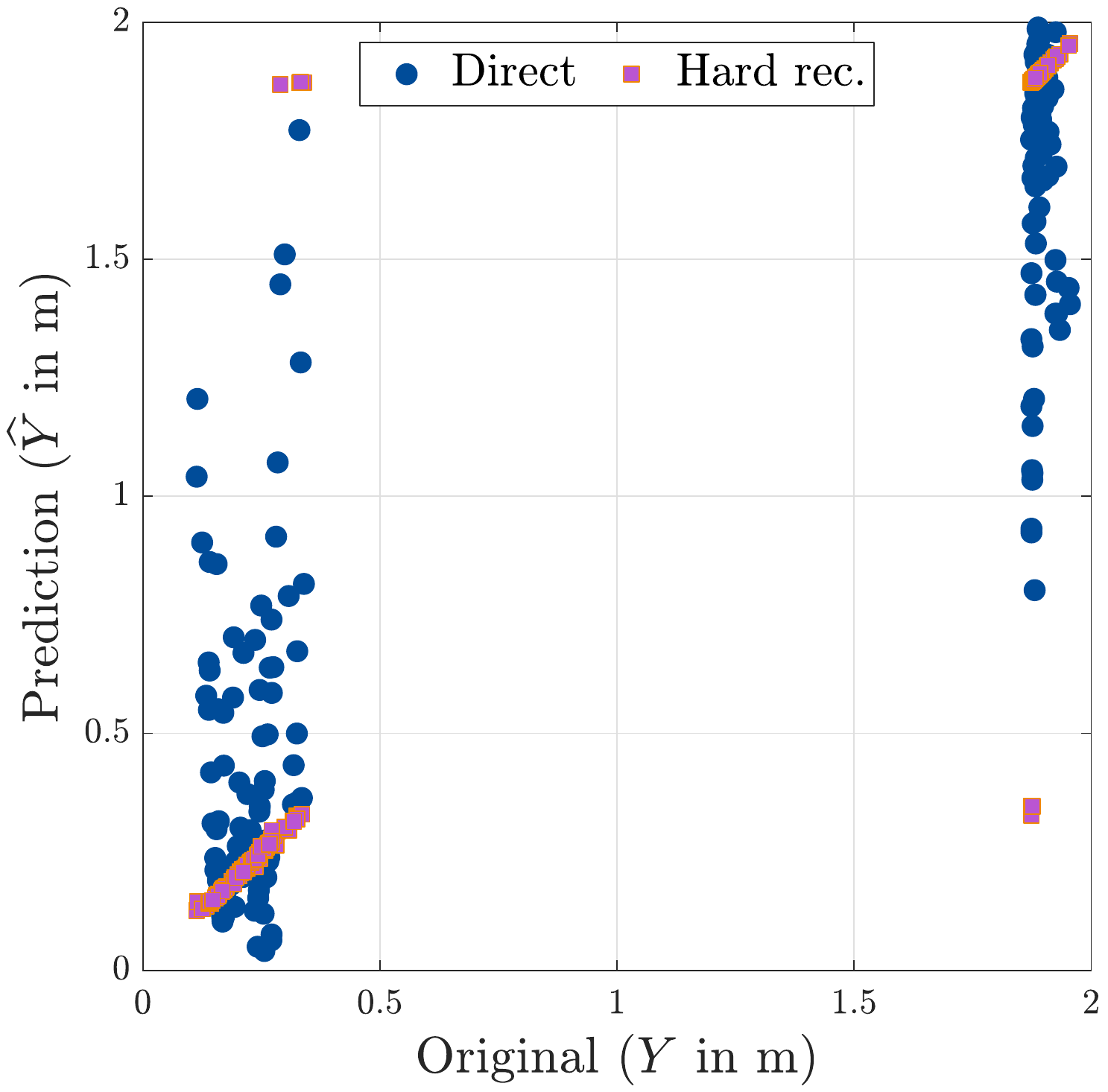}}%
	\subfloat[Soft reconstruction and categorical Kriging]{\label{fig:Ex2:YY_b}\includegraphics[width=0.45\textwidth]{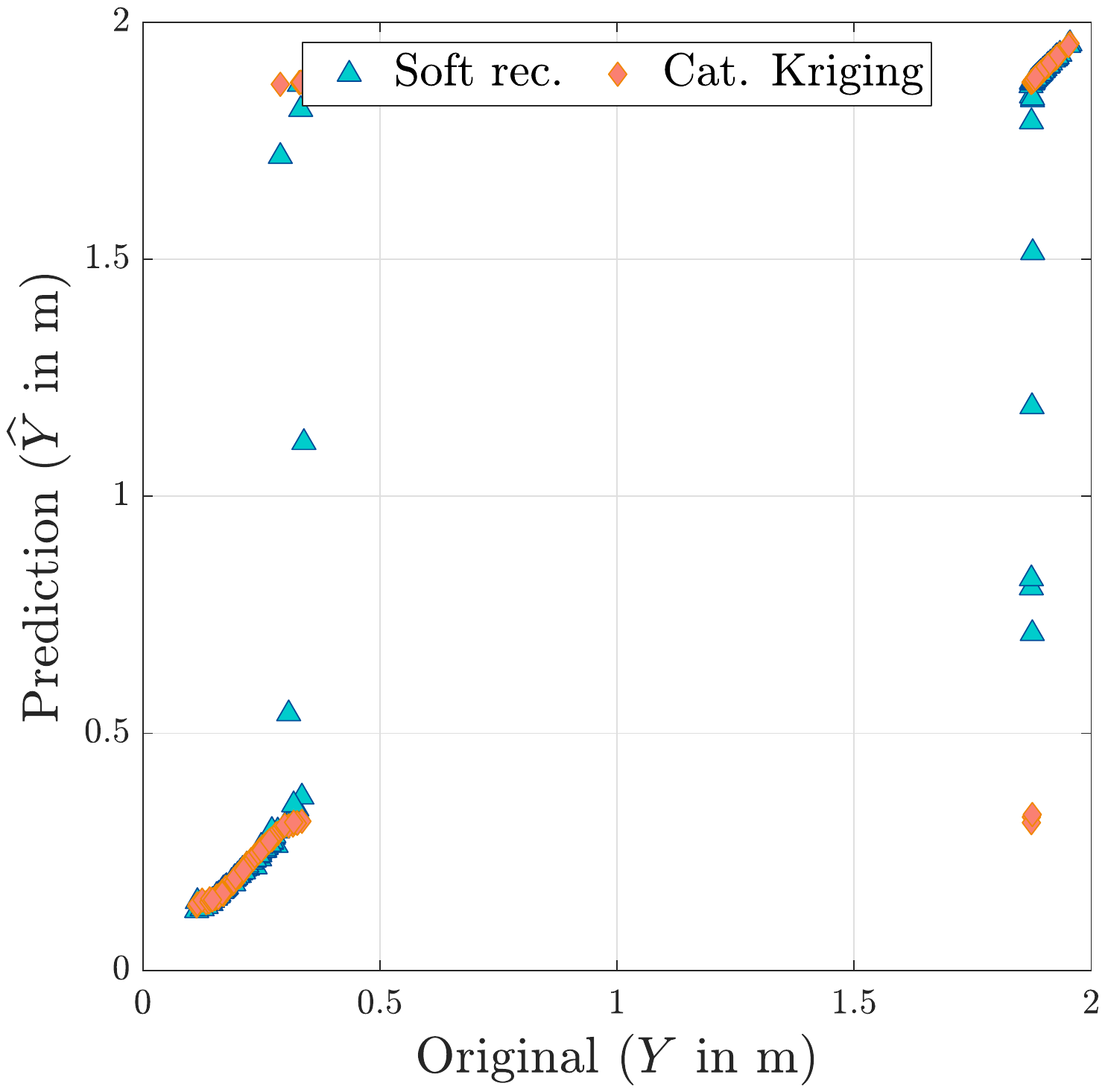}}%
	\caption{Example 2: Original  \emph{vs.} predicted vertical displacement for different approximation techniques.}
	\label{fig:Ex2:YY}
\end{figure}

We run the analysis using the proposed method and considering three different experimental design sizes and $20$ repetitions. The resulting errors are summarized as boxplots shown in Figure~\ref{fig:Ex2:Results}. The first observation is that the difference between the results obtained by the proposed method and a direct Kriging model (\ie a single Kriging model built using the entire data set) is much more important than in the previous case, often by orders of magnitude. This is due to the fact that the two regimes of non-linear structure behaviours are prominently different as shown in Figure~\ref{fig:Ex2:YY}.  Furthermore in this example, categorical Kriging performs quite well. It is not clear however which recombination approach is the best. When looking at the normalized mean square error, the hard recombination is slightly better. This is the opposite when looking at the mean absolute error, \emph{i.e.}, the soft and categorical recombination are slightly better. 

Figure~\ref{fig:Ex2:YY} shows the original \emph{vs.} predicted vertical displacement for the four approximations using a random subset of the validation set of size $200$. The left panel of this figure shows how a single model (called "direct") spans the entire range between the two regimes of the truss and leads to huge errors. In contrast, the multi-stage approaches properly detect the discontinuities. It is also clear from this figure how the recombination scheme affects the final prediction when there are classification errors. The soft recombination reduces the error for those cases when there is uncertainty in the classification. Note that the same outlier points are observed in Figures~\ref{fig:Ex2:YY_a}~and~\ref{fig:Ex2:YY_b} when hard reconstruction and categorical Kriging are used: these outliers only stem from classification error. 

\subsection{Tensile fabric structure}
In this final example, we investigate a model that simulates the behaviour of a tensile membrane structure (TMS) under extreme loading \citep{Valdes-Vazquez2020,Valdes-Vazquez2021}. TMS are flexible lightweight structures made of composite fabric spanning long distances. They have many advantages in terms of architectural sophistication but are yet challenging to design. By their very nature, they are unable to carry out-of-plane moments and shear forces that may result from the extreme wind loads they are expected to withstand. They further require careful pre-stressing to keep a stable form. 

\begin{figure}[!ht]
	\centering
	\subfloat[Top view]{\label{fig:Ex3:hypar_a}\includegraphics[width=0.45\textwidth]{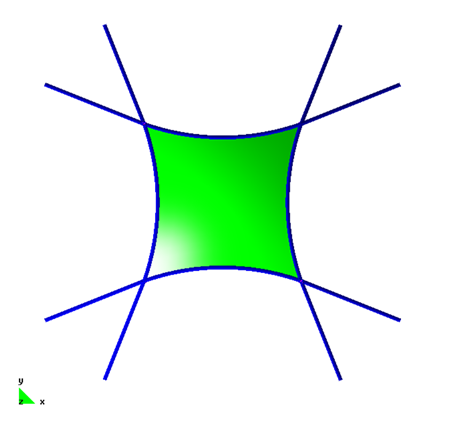}}%
	\subfloat[Side view]{\label{fig:Ex3:hypar_b}\includegraphics[width=0.45\textwidth]{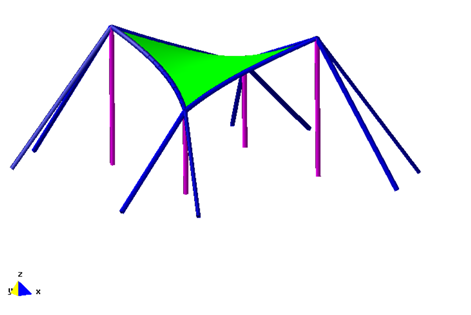}}%
	\caption{Hypar structure considered in this study.}
	\label{fig:Ex3:hypar}
\end{figure}

\begin{figure}[!ht]
	\centering
	\includegraphics[width=0.45\textwidth]{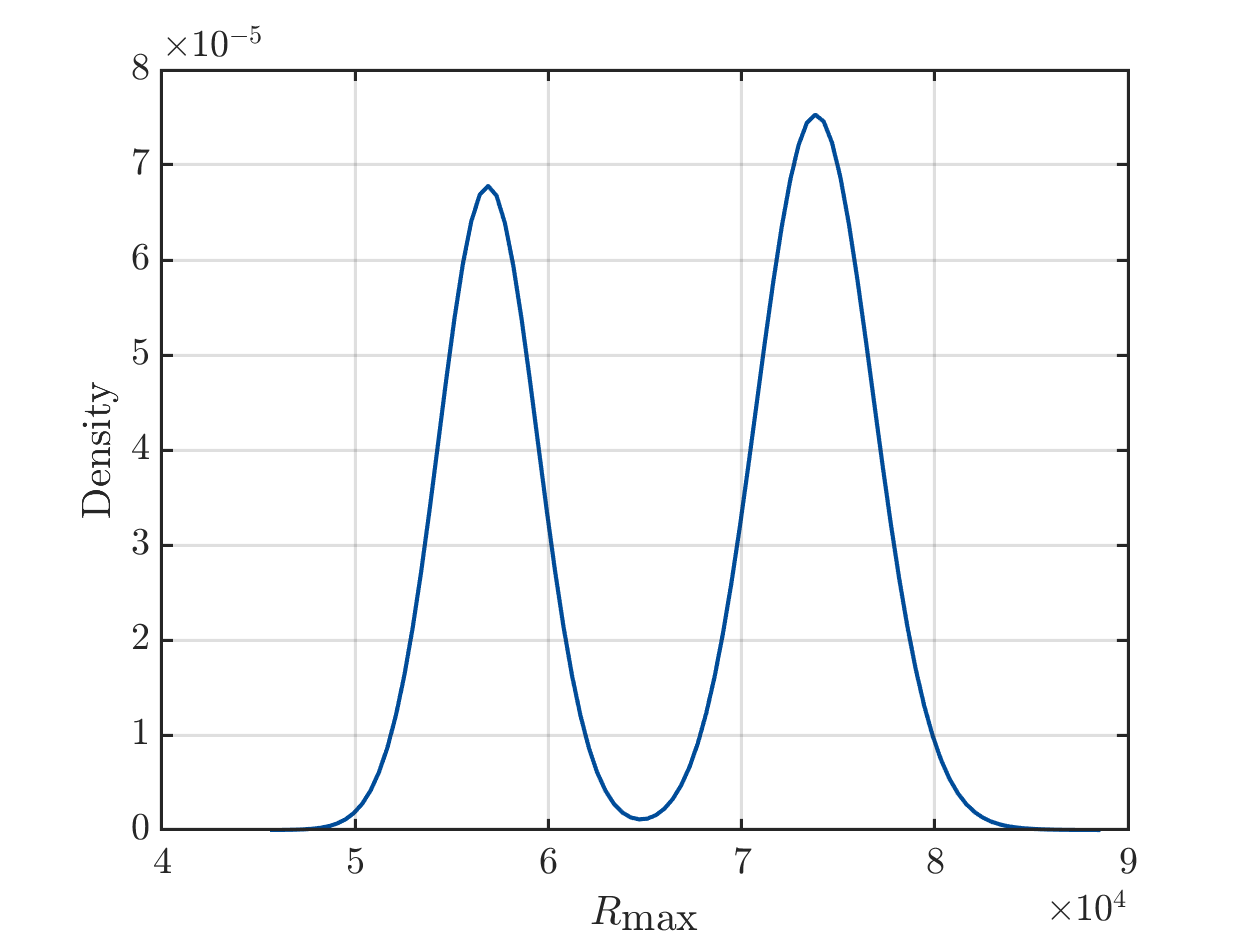}%
	\caption{Example 3: Kernel smoothing density of the maximum reaction force of the hypar.}
	\label{fig:Ex3:PDF}
\end{figure}

Special codes are designed to simulate the response of complex tensile membrane structures. \textsc{Comet} is one such in-house finite element code developed at the University of Gua \citep{Valdes-Vazquez2021}. In this work, we consider a hypar (hyperbol-paraboloid), which is one of the most common shapes for TMS, designed using \textsc{Comet} and illustrated in Figure~\ref{fig:Ex3:hypar}. The probabilistic model is described using the random variables presented in Table~\ref{tab:example3}. There are various quantities of interest for such a design model. We consider here the maximum reaction forces on the supports of the system (cables or mast). It turns out that according to the boundary conditions, the maximum reaction force occurs in two different locations with entirely different magnitudes. This is shown by the bi-modality of the kernel density estimate of the model response in Figure~\ref{fig:Ex3:PDF}. 
\begin{table}[!ht]
	\begin{center}
		\begin{tabular}{*{4}{l}}
			\hline
			Parameter & Distribution & Mean & C.o.V. \\
			\hline
			Wind load ($V_w$ - m/s) & Gumbel & $36.11$ & $0.132$ \\
			Cable pre-stress ($S_{xx}$ - N/m$^2$)  & Gaussian & $5.09 \cdot 10^8$ & $0.06$ \\
			Young's modulus ($E_{wf}$ - N/m)  & Lognormal & $8 \cdot 10^5$ & $0.07$ \\
			Poisson modulus ($\nu$ - ) & Gaussian & $0.4$ & $0.05$ \\
			Fabric prestress warp ($F_w$ - N/m$^2$) & Gaussian & $4 \cdot 10^6$ & $0.05$ \\
			Fabric prestress fill ($F_f$ - N/m$^2$) & Gaussian & $4 \cdot 10^6$ & $0.05$ \\
			Mast Young's modulus ($E_m$ - N/m$^2$) & Lognormal & $2.1 \cdot 10^{11}$ & $0.03$ \\
			Cables Young's modulus ($E_c$ - N/m$^2$) & Lognormal & $2.1 \cdot 10^{11}$ & $0.03$ \\
			Mast cross-sectional area ($A_m$ - m$^2$) & Gaussian & $1.7 \cdot 10^{-3}$ & $0.032$ \\
			Cable cross-sectional area  ($A_c$ - m$^2$) & Gaussian & $7.854 \cdot 10^{-5}$ & $0.032$ \\ 
			\hline
		\end{tabular}
	\end{center} 
	\caption{Hypar structure: probabilistic input model.}
	\label{tab:example3}
\end{table}

The underlying mechanisms leading to each of two model response modes are different and building a single surrogate model to account for both leads to inaccurate results. We consider then the three-stage approach proposed in this paper, with an experimental design of size $500$ and a validation set of size $1,000$. The experimental design is split into five different subsets of sizes $100$, $200$, $300$, $400$ and $500$. In each of these, the DPMM clustering rightly identifies that there are two sets of responses.

\begin{figure}[!ht]
	\centering
	\subfloat[$NMSE$]{\label{fig:Ex3:Results_a}\includegraphics[width=0.45\textwidth]{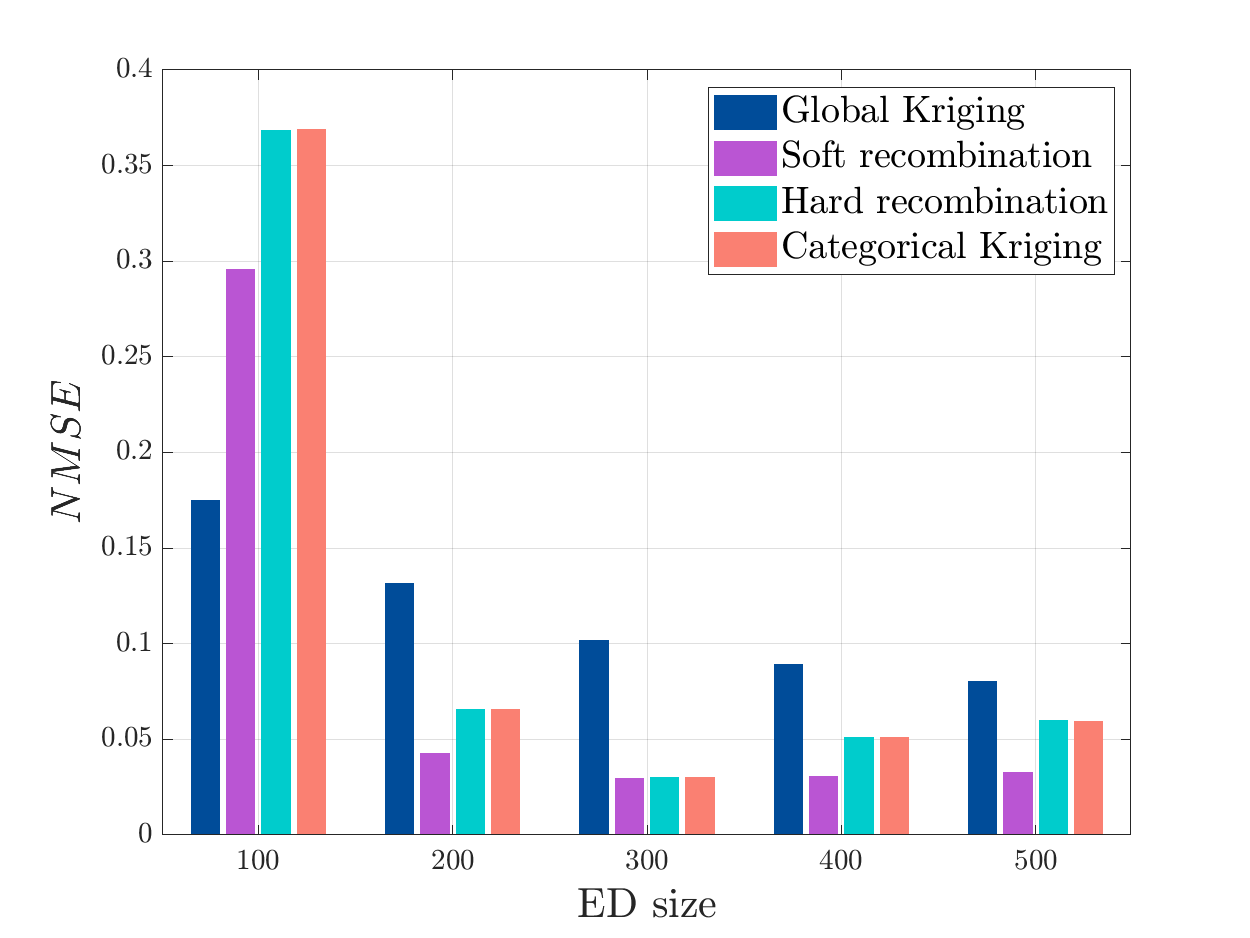}}%
	\subfloat[$MAE$]{\label{fig:Ex3:Results_b}\includegraphics[width=0.45\textwidth]{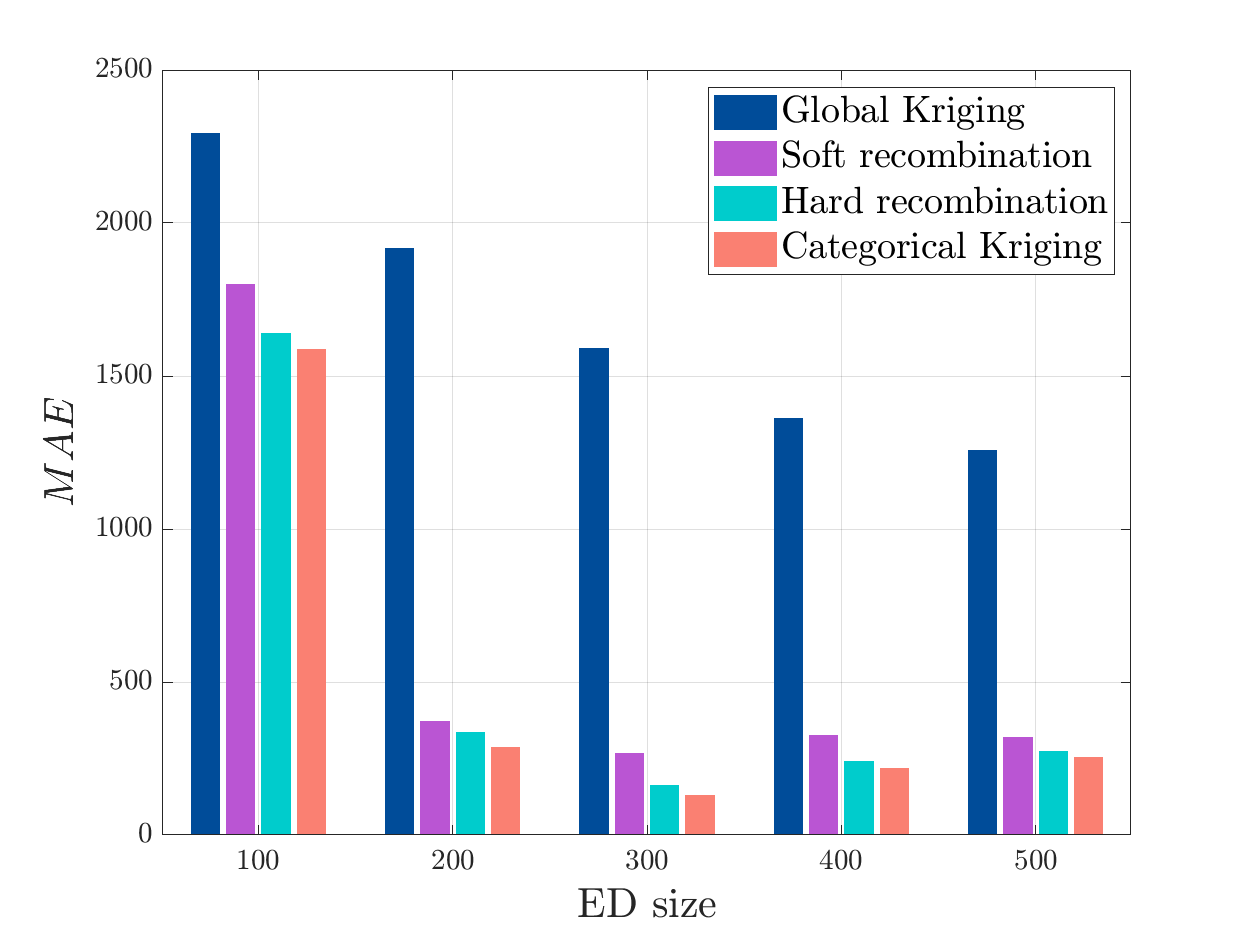}}%
	\caption{Example 3: Computed errors for the hypar structure for increasing experimental design sizes.}
	\label{fig:Ex3:Results}
\end{figure}

Figure~\ref{fig:Ex3:Results} shows the resulting NMSE and MAE for each experimental design size. As expected, the error decreases with increasing ED size and our proposed workflow yields more accurate approximations than a global single Kriging model, except for NMSE when $N=100$ due to the large weight of misclassification errors. The soft recombination is slightly better than hard recombination and categorical Kriging which have very similar predictions. 

\begin{figure}[!ht]
	\centering
	\subfloat[$N=100$ - Direct and soft recombination]{\label{fig:Ex3:Density_a}\includegraphics[width=0.45\textwidth]{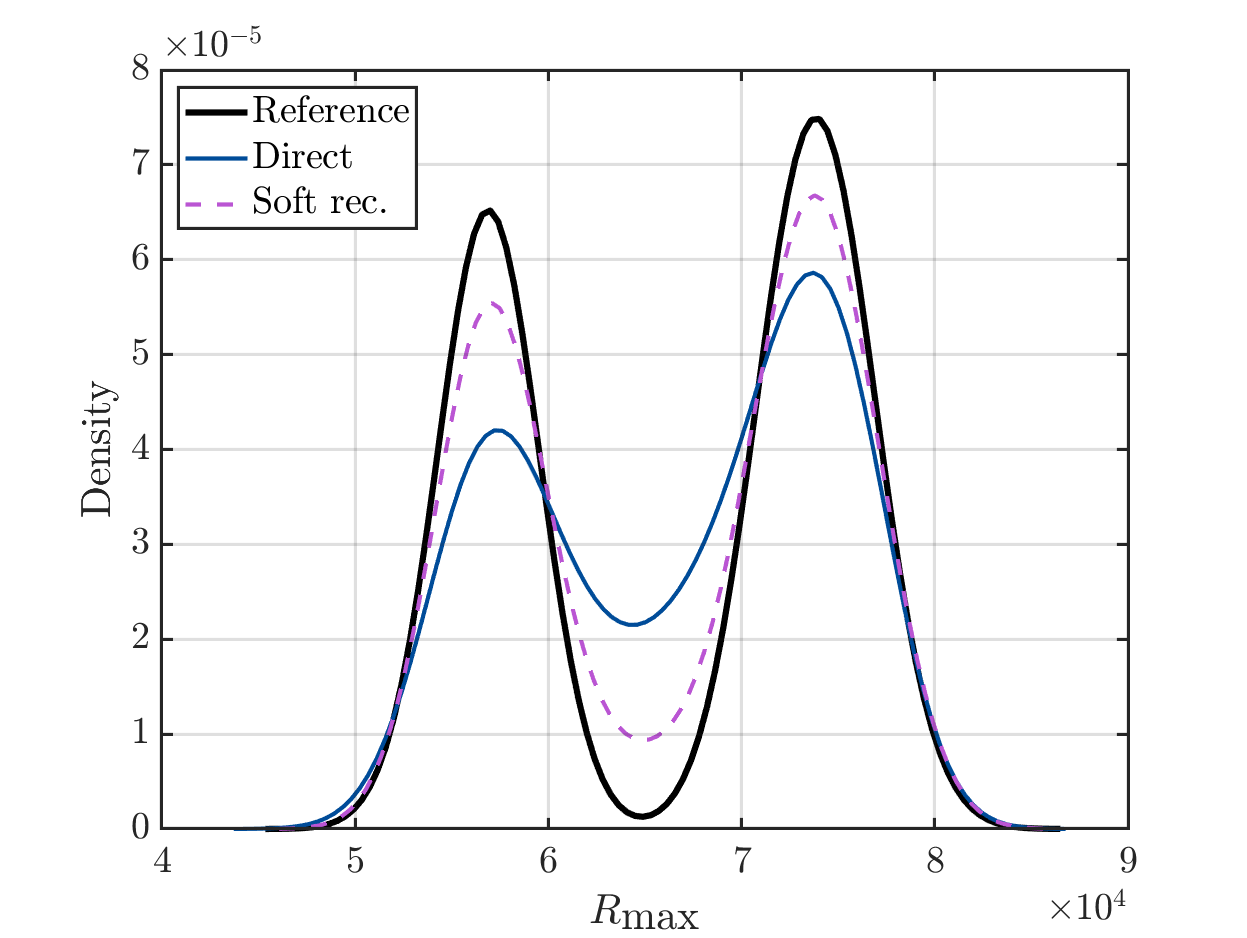}}%
	\subfloat[$N=100$ - Hard recombination and categorical Kriging]{\label{fig:Ex3:Density_b}\includegraphics[width=0.45\textwidth]{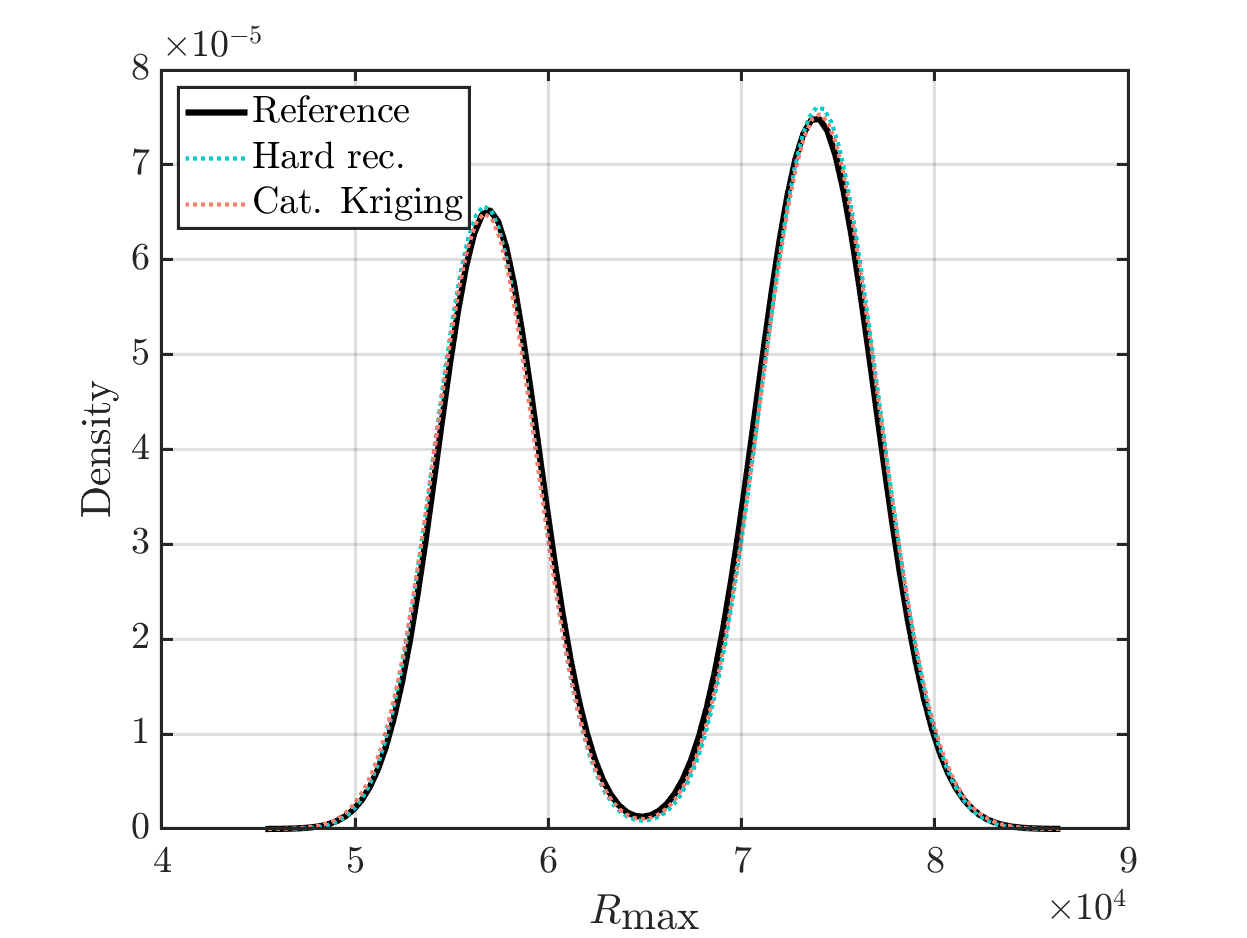}}%
	\\
	\subfloat[$N=300$ - Direct and soft recombination]{\label{fig:Ex3:Density_c}\includegraphics[width=0.45\textwidth]{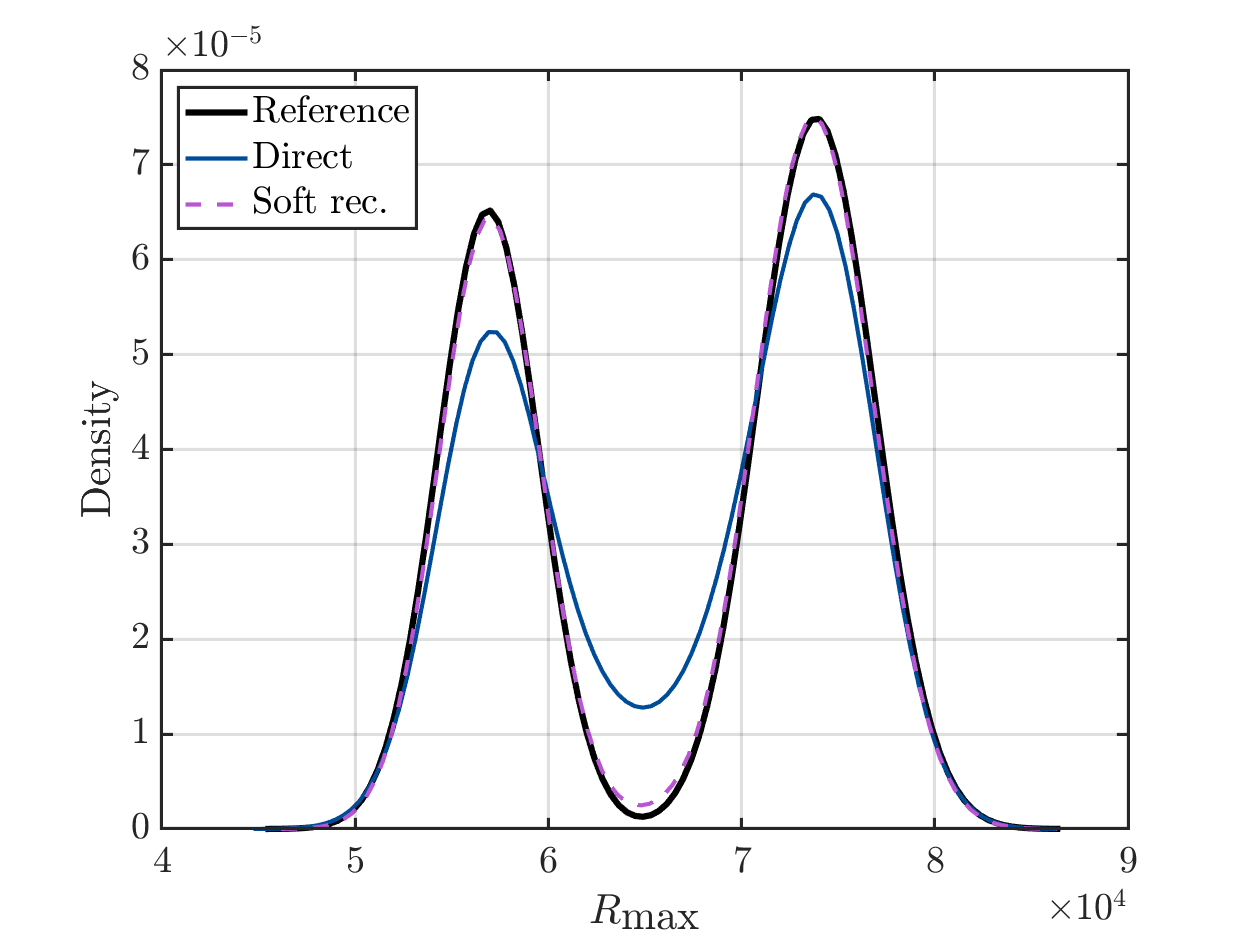}}%
	\subfloat[$N=300$ - Hard recombination and categorical Kriging]{\label{fig:Ex3:Density_d}\includegraphics[width=0.45\textwidth]{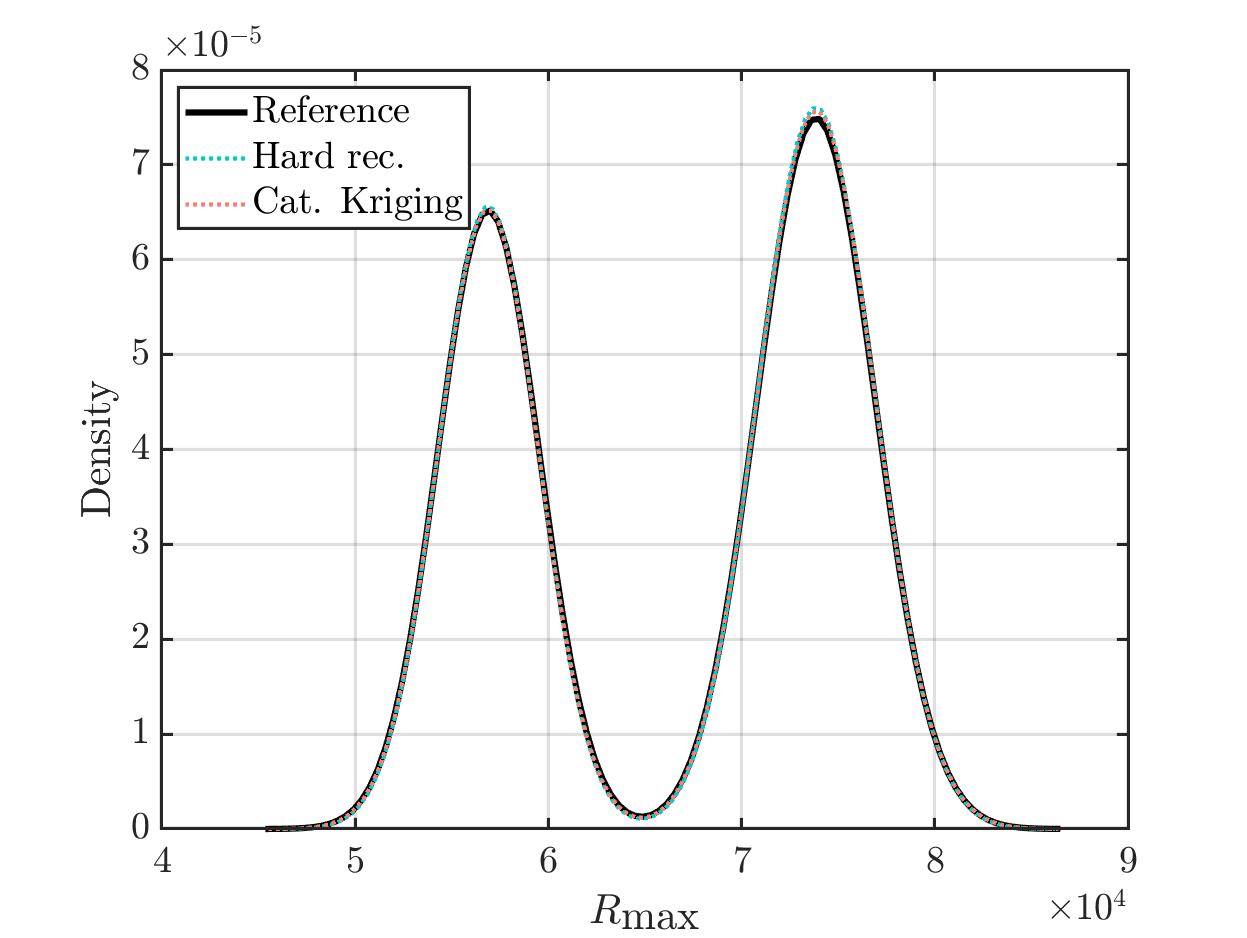}}%
	\caption{Example 3: Computed errors for the hypar structure for increasing experimental design sizes.}
	\label{fig:Ex3:Density}
\end{figure}

Finally, Figure~\ref{fig:Ex3:Density} shows PDFs of the responses for different models with ED sizes of $100$ and $300$. We can see that even for $100$ samples, the densities with the hard recombinations are extremely similar to those obtained from the original model. This shows that the reconstructed surrogate models are extremely accurate except for a few outliers which are due to misclassification in the second step of the workflow. The soft recombination puts more mass in the middle of the density support, due to the weighted recombination. This mass reduces as the ED size increases. 
\section{Conclusion}
Surrogate modelling is now a well-established method that allows one to reduce the computational burden of simulation intensive methods that require multiple evaluations of a costly computational model. Building an accurate surrogate model with limited data generally requires that the functions to approximate are smooth and regular. This is however not always the case in many applications, \eg crash simulation or computational fluid dynamics. 

In this paper, we propose a three-stage approach for the approximation of non-smooth functions for systems exhibiting multiple behaviours and/or discontinuities. The problem is tackled by dividing the task into three complementary parts: i. a joint input-output clustering stage that identifies the different patterns exhibited by the system using a non-parametric Bayesian approach, namely a Dirichlet process mixture model, ii. a partition of the input space according to the identified clusters using support vector machines, and eventually iii. the construction of local surrogates, herein Kriging models, using data from each of the partitions. For any new point, the prediction is made by appropriately recombining the predictions made by each of the Kriging models, according to the assigned class of the new point.

The proposed approach is validated on two analytical examples and an engineering application (FE-based tensile membrane structure). It is shown to be both accurate and efficient compared to a traditional surrogate modelling approach ignoring the non-smoothness.  

The three methods selected for each stage all provide probabilistic predictions. While the posterior probabilities of the support vector machines classifiers have been used within the soft reconstruction scheme, the ones provided by the Dirichlet process mixture models have not been exploited yet. However, as seen in the examples, mislabelling the initial data leads to large errors. These could be reduced by accounting for the uncertainties in the clustering stage. In a future work, we intend to account for the latter so as to provide a fully probabilistic prediction scheme that propagates the epistemic uncertainties from one step to the next. 

\newpage
\bibliographystyle{chicago}
\bibliography{References}
\end{document}